\documentclass[runningheads]{llncs}

\usepackage{eccv}

\usepackage{placeins}

\usepackage{pgfplots}
\pgfplotsset{compat=1.18}
\usepackage{pgf-pie}

\usepackage{eccvabbrv}

\usepackage{graphicx}
\usepackage{booktabs}
\usepackage{url}
\usepackage{xcolor}
\usepackage{wrapfig}
\usepackage{multirow}
\usepackage{colortbl}
\usepackage{gradient-text}
\usepackage{tcolorbox}
\usepackage{bookmark}

\usepackage[accsupp]{axessibility}  %

\definecolor{darkgreen}{RGB}{0,140,60}
\definecolor{almond}{rgb}{0.94, 0.87, 0.8}

\usepackage{orcidlink}

\begin{document}

\title{AVRT: Audio-Visual Reasoning Transfer through Single-Modality Teachers}

\titlerunning{Audio-Visual Reasoning Transfer}

\author{Edson Araujo\inst{1} \and
Saurabhchand Bhati\inst{2} \and
M. Jehanzeb Mirza\inst{2} \and
Brian Kingsbury\inst{3} \and
Samuel Thomas\inst{3} \and
Rogerio Feris\inst{3,4} \and
James R. Glass\inst{2} \and
Hilde Kuehne\inst{4,5}}

\authorrunning{Araujo et al.}

\institute{University of Tübingen, Germany \and
MIT, Cambridge MA, USA \and
IBM Research, USA \and
MIT-IBM Watson AI Lab, USA \and
Tuebingen AI Center, Germany}

\maketitle

\renewcommand{\thefootnote}{\dag}
\begin{abstract}
	Recent advances in reasoning models have shown remarkable progress in text-based domains, but transferring those capabilities to multimodal settings, e.g., to allow reasoning over audio-visual data, still remains a challenge, i.a., because of the limited availability of high-quality reasoning data in targeted multimodal combinations.
    To address this problem, we introduce AVRT, a novel framework that generates high-quality audio-visual reasoning traces from single-modality teacher models.
    We generate independent vision- and audio-reasoning traces via models specialized to reason over their respective modalities and merge the resulting traces with an LLM merger model. 
    The resulting multimodal traces are used in a supervised fine-tuning (SFT) cold start to adapt the target model to audio-visual reasoning traces first, before training it in a second reinforcement learning stage on larger-scale data.
    Evaluated on seven audio-visual and audio benchmarks, our 3B and 7B parameter models achieve state-of-the-art results among models of comparable size including OmniBench and DailyOmni for audio-visual and MMAR for audio-only reasoning, showing that cross-modal training also transfers to single-modality tasks and establishing a new training pipeline for multimodal reasoning models. \footnote{All code, data, and checkpoints will be made available.}
\end{abstract}

\section{Introduction}
Humans perceive the world by combining information from multiple modalities through diverse sensory inputs.
With the wide availability of multimodal data, such as videos, multimodal understanding in general and audio-visual understanding in particular has drawn more and more interest from the research community. 
Recent advancements in this area, also in combination with large language models, have shown remarkable performance in audio-visual understanding ~\cite{damonlpsg2024videollama2,Qwen2.5-Omni,liu2025ola,gemini2.5,gpt4o}. 

In parallel, the emergence of reasoning-capable language models has led to new  capabilities with respect to the analysis and understanding of a given scenario, exemplified by OpenAI's o-series~\cite{jaech2024openai} and DeepSeek-R1~\cite{deepseekai2025deepseekr1incentivizingreasoningcapability}. 
These advances have been significantly driven by reinforcement learning techniques~\cite{shao2024deepseekmath}.
These reasoning capabilities have been successfully extended to inputs such as vision-text models~\cite{huang2025visionr1,dong2025insight} and audio-text models~\cite{xie2025audio,wen2025sari,goel2025audio}, demonstrating chain-of-thought capabilities within the respective modalities. 
However, audio-visual reasoning has not yet reached the same level of advancement as its single-modality counterparts, i.a., due to the challenge of integrating information and reasoning cues across different modalities at scale, as well as due to the practical lack of audio-visual reasoning data in general.
Existing approaches try to address this problem, e.g., by generating reference reasoning chains from large foundation teacher models that were trained with all target modalities~\cite{du2025crab} or by trying to approach the problem by extending reinforcement learning formulations, e.g., by improved credits assignment or by context summarization ~\cite{yang2025humanomniv2,kulkarni2025avatarreinforcementlearningsee}.
\begin{figure*}[t]
    \centering
    \includegraphics[width=\textwidth]{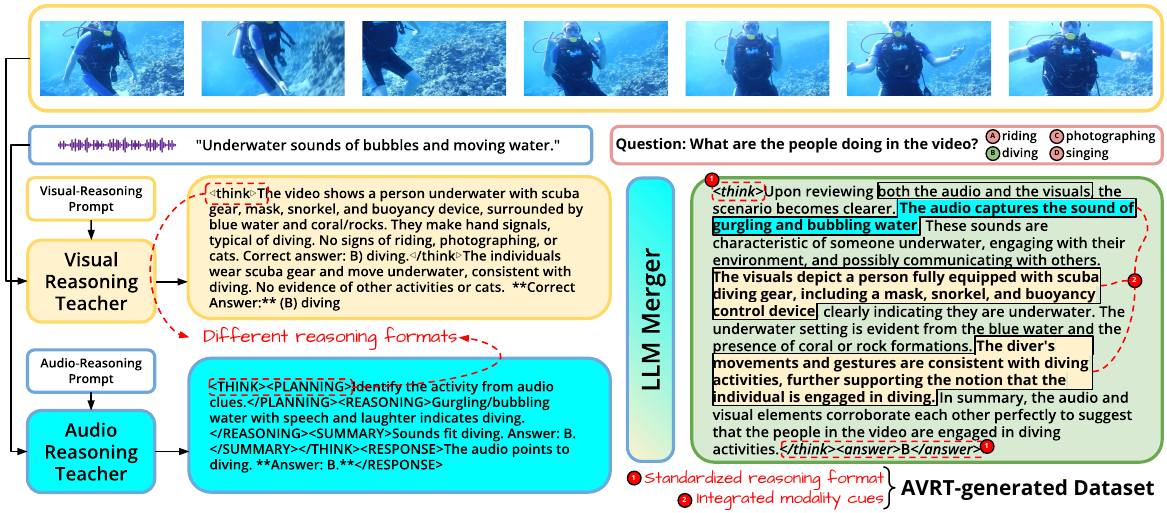}
    
    \caption{\textbf{Overview of the AVRT pipeline:} We first generate reasoning chains from single-modality teacher models that are prompted in the format they were optimized for and, second, leverage an LLM merger as an interface between the teacher models and the resulting reasoning chain to aggregate the information and put it into the target format. The resulting audio-visual traces are then used to train a student model in a cold-start manner via supervised fine-tuning (SFT) followed by GRPO fine-tuning for audio-visual question-answering.}
    \label{fig:teaser}
    
\end{figure*}

This paper proposes a new pipeline for Audio-Visual Reasoning Transfer (AVRT) based on single-modality teachers. To this end, AVRT extracts detailed chains of thought independently from specialized visual and audio teachers, then merges them with a text-only aggregator into single cross-modal reasoning traces. An overview is shown in Fig.~\ref{fig:teaser}.
Namely, we combine traces from specialized single-modality reasoning models via a text-only LLM merger model to generate a coherent multimodal reasoning that explicitly correlates information across audio and visual channels.
By using an LLM as a merging interface for the different teacher models, every model can be prompted in the format that it was trained for, leading to high-quality modality-specific outputs. The following LLM merger then allows to aggregate the information, including different meta tags, and to convert it into a consistent target format.
The resulting audio-visual traces are then in the first step used to fine-tune a student audio-visual LLM in a cold-start manner via supervised fine-tuning (SFT) to learn the reasoning formatting and patterns, as well as to integrate multimodal reasoning into the student model. 
In a second step, the student model is trained following the GRPO~\cite{shao2024deepseekmath} objective.

We evaluate the proposed approach on seven challenging datasets: First, as zero-shot vision-audio downstream datasets, we consider DailyOmni~\cite{zhou2025dailyomniaudiovisualreasoningtemporal}, OmniBench~\cite{li2024omnibench}, Video-MME~\cite{fu2025video}, and RivaBench~\cite{sun2025video}; second, we evaluate on AVQA~\cite{yang2022avqa} as in-domain validation dataset; and finally on MMAR~\cite{ma2025mmar} and MMAU~\cite{sakshi2024mmau} as audio-only downstream task.
Considering a 3B-parameter and a 7B-parameter student models, we achieve state-of-the-art performance compared to other 3B and 7B audio-visual reasoning models.
Our ablation shows that the combination of reasoning traces of two different teacher models by a language-only merger model improves audio-visual reasoning in the GRPO learning stage.
Our contribution can be summarized as follows:
1) We propose a novel method that aggregates single-modality reasoning traces into coherent multimodal reasoning via a text-only merger, a recipe applicable to any modality combination lacking reasoning data, which we instantiate here for the audio-visual domain.
2) We are the first to train a multimodal reasoner based on composed single-modality reasoning data, demonstrating that cross-modal reasoning can emerge from composed single-modality expertise without requiring a natively multimodal reasoner as teacher.
3) We conduct an extensive evaluation on audio-visual benchmarks, achieving state-of-the-art results that compete with larger models. We show that single-modality teachers composed through a text-only merger outperform what each teacher achieves individually (see Table~\ref{tab:ablation_reasoning_chains}), and that training on audio-visual data improves single-modality performance (see Table~\ref{tab:ablation_single_modality}), providing evidence of genuine cross-modal reasoning transfer.

\section{Related Work}
\label{sec:rw}
\textbf{Audio-Visual Large Language Models.} Audio-visual large language models have made constant progress in addressing challenging tasks in multimodal understanding. Early works like Meerkat \cite{chowdhury2024meerkat} focus on fine-grained spatial and temporal grounding on five audio-visual tasks, introducing optimal transport-based modality alignment and cross-attention modules for audio-visual consistency. AVicuna \cite{tang2025empowering} proposes specifically targeting temporal referential dialogue in untrimmed videos, introducing the Audio-Visual Tokens Interleaver for temporal alignment. VideoLLaMA 2 \cite{damonlpsg2024videollama2} advances spatial-temporal modeling through specialized Spatial-Temporal Convolution connectors and enhanced audio understanding via joint training, achieving state-of-the-art performance among open-source models. Rather than developing new models from scratch, PAVE \cite{liu2025pave} introduces a lightweight adaptation framework that extends existing video LLMs to other modalities through efficient ``patches'' that add only 0.1\% additional parameters. More recent work moves toward comprehensive omni-modal capabilities, with Qwen2.5-Omni \cite{Qwen2.5-Omni} enabling end-to-end streaming multimodal inputs and outputs through innovations like TMRoPE for synchronizing video with audio and the Thinker-Talker architecture for concurrent text and speech generation. Ola \cite{liu2025ola} proposes progressive modality alignment training strategies that use video as a central bridge to connect modalities.
Despite these advances, these existing approaches often struggle to effectively associate information across both modalities, lacking structured reasoning processes that can explicitly capture and use cross-modal dependencies and correlations for comprehensive multimodal understanding. AVRT addresses this gap by introducing explicit cross-modal reasoning traces generated from single-modality specialists, enabling structured reasoning that correlates information across audio and visual channels without requiring end-to-end multimodal pretraining.

\textbf{Audio-Visual Reasoning.} Audio-visual reasoning with large language models has seen rapid advancements. Daily-Omni \cite{zhou2025dailyomniaudiovisualreasoningtemporal} introduces a dedicated Audio-Visual QA dataset with an agent leveraging visual, audio, and speech modalities. Recent work has increasingly leveraged reinforcement learning: EchoInk-R1 \cite{xing2025echoinkr1exploringaudiovisualreasoning} uses GRPO for structured cross-modal reasoning, HumanOmniV2 \cite{yang2025humanomniv2} addresses shortcut problems through context summarization and logical rewards, Omni-R1 \cite{zhong2025omnir1reinforcementlearningomnimodal} handles temporal-spatial trade-offs, AVATAR \cite{kulkarni2025avatarreinforcementlearningsee} introduces Temporal Advantage Shaping for credit assignment, and Video-RTS \cite{wang2025video} combines data-efficient RL with test-time scaling. video-SALMONN-o1 \cite{sun2025video} is the first open-source reasoning-enhanced audio-visual LLM with pDPO training and the RivaBench benchmark. Building on the success of Qwen2.5-Omni, Qwen3-Omni \cite{xu2025qwen3} extends omni-modal capabilities to reasoning with its MoE architecture. More recently, AURELIA \cite{chowdhury2025aurelia} introduces a test-time reasoning distillation approach that uses three different forward passes to generate step-by-step reasoning, which is then provided as input alongside the audio-visual data and question to guide the model's reasoning process. 
Rather than relying on test-time adaptation strategies that require multiple inference passes we use single-modality specialist models as teachers to enhance a multimodal student model through knowledge distillation, constructing a dataset with explicit audio-visual reasoning chains. This allows for a supervised fine-tuning (SFT) approach, supplemented by RL, to produce a model that generates answers in a single forward pass, significantly reducing inference cost compared to multi-pass test-time strategies.

\begin{table*}[t]
    \centering
    \resizebox{\textwidth}{!}{%
    \begin{tabular}{l c c c}
    \toprule
    \textbf{Dataset} & \textbf{Modalities} & \textbf{Answer Format} & \textbf{\# QA pairs (K)} \\
    \midrule
    AVQA & A+V & Video + question + 4-way answer & 57.3  \\
    OmniInstruct-v1 & A+I & Image + question + 4-way answer & 96.1  \\
    MUSIC-AVQA & A+V & Video + question + 4-way answer (focused on music) & 45.9 \\
    AVQA\,-\,R1\,-\,6K & A+I & Video + question + 4-way answer (subset of OmniInstruct) & 6.4  \\
    \multicolumn{1}{>{\columncolor{almond}}l}{} & \multicolumn{1}{>{\columncolor{almond}}c}{} & \multicolumn{1}{>{\columncolor{almond}}c}{Video + question + 4-way answer + Reasoning chains} & \multicolumn{1}{>{\columncolor{almond}}c}{}  \\
    \multicolumn{1}{>{\columncolor{almond}}l}{\multirow{-2}{*}{\parbox{2.5cm}{\textbf{AVRT-20K }\scriptsize{ (ours)}}}} & \multicolumn{1}{>{\columncolor{almond}}c}{\multirow{-2}{*}{A+V}} & \multicolumn{1}{>{\columncolor{almond}}c}{incorporating audio and visual data} & \multicolumn{1}{>{\columncolor{almond}}c}{\multirow{-2}{*}{19.2}}  \\
    \bottomrule
    \end{tabular}%
    }
    
    \caption{\textbf{Comparison of audio-visual question answering datasets.} AVRT-20K (derived from AVQA) provides reasoning traces that explicitly integrate audio (A) and visual (V) modalities, addressing a key limitation in existing AVQA datasets which focus solely on question-answer pairs without intermediate reasoning steps. All datasets use multiple-choice questions with 4 options (MCQ-4). The number of QA pairs is reported in thousands (K).}
    \label{tab:datasets_unified}
    
\end{table*}

\textbf{Audio-Visual Datasets.} AVQA \cite{yang2022avqa} can be considered one of the foundational audio-visual QA datsets with $57,335$ question-answer pairs from daily audio-visual activities requiring clues from both modalities. OmniInstruct \cite{li2024omnibench} develops a comprehensive tri-modal reasoning dataset combining visual, audio, and textual resources, while MUSIC-AVQA \cite{Li2022Learning} expands to musical performance with $45,867$ question-answer pairs across $9,288$ videos. AVQA-R1-6K \cite{xing2025echoinkr1exploringaudiovisualreasoning} provided a manually curated subset of OmniInstruct focusing on questions that are more likely to require audio-visual reasoning. While these datasets have advanced the field significantly, they primarily focus on question-answer pairs without providing explicit reasoning traces that demonstrate how models should integrate cross-modal information. Our AVRT approach directly addresses this gap through structured reasoning chain generation, as compared in Table~\ref{tab:datasets_unified}. By providing explicit cross-modal reasoning traces rather than answer-only supervision, AVRT enables models to to integrate audio and visual evidence.

\section{Methodology}

In this paper, we derive audio-visual reasoning traces based on audio-visual question-answering pairs as e.g. provided by AVQA \cite{yang2022avqa}.
We first discuss the generation of audio-visual reasoning traces in Sec. \ref{sec:dataset} and our training procedure to leverage this data to perform audio-visual question answering in Sec. \ref{sec:training}.
\subsection{Cross-Modal Reasoning Trace Generation}
\label{sec:dataset}
Our data generation process consists of two main stages: a single-modality reasoning extraction step and a cross-modal aggregation step. Let an audio-visual question-answering data sample be denoted as $(X, Q)$, where $X$ is a video and $Q$ is a question. The video $X$ comprises both an audio stream $A$ and a visual stream $V$, such that $X = (A, V)$.

\noindent\textbf{Single-Modality Reasoning Extraction.} First, we generate modality-specific reasoning. We select specialized teacher models for the audio ($T_A$) and visual ($T_V$) modalities. For a given sample $(X,Q)$, we provide each teacher with the question and its corresponding modality. We use carefully crafted prompts, $P_A$ and $P_V$, to elicit detailed reasoning traces. The audio reasoning trace is generated as $R_A = T_A(Q, A, P_A)$, and the visual reasoning trace is $R_V = T_V(Q, V, P_V)$. These traces capture the unique characteristics and patterns of each modality.

\noindent\textbf{Cross-Modal Aggregation.} In the second stage, we perform cross-modal aggregation. We use a text-only large language model, $M_{agg}$, to merge the reasoning outputs. This model takes the reasoning traces from both modalities and the original question to produce a unified, cross-modal reasoning output: $R_{agg} = M_{agg}(Q, R_A, R_V)$. This aggregation step transforms the diverse reasoning formats into a uniform structure, correlating characteristics from both modalities and incorporating cross-modal dependencies. 

\noindent\textbf{Filtering.} %
To avoid introducing noise during the SFT, we consider only reasoning traces where both modality-specific teachers generate correct responses. This filtering strategy ensures high-quality training data by avoiding the propagation of erroneous reasoning patterns that could introduce noise during cross-modal aggregation \cite{turpin2023language, xie2024order}. As a result, we only keep a 20K subset of the original 40K AVQA data samples. An ablation on the filtering strategy is provided in Section~\ref{sec:ablation}, showing that training on filtered data does not affect generalization, but also showing that training on unfiltered data only slightly decreases performance leaving room for scaling on unlabeled data. 

\subsection{Training} \label{sec:training}
\textbf{Stage 1: Supervised Fine-Tuning.}
We fine-tune the base model on the merged audio-visual reasoning traces using an autoregressive language loss. Given a training sample $(X, Q, R_{agg})$ where $X = (A, V)$ is the video with audio and visual streams, $Q$ is the question, and $R_{agg}$ is the aggregated reasoning trace, we optimize the cross-entropy loss:
  \begin{equation}
  \mathcal{L}_{SFT} = -\sum_{t=1}^{|R_{agg}|} \log p_\theta(r_t | X, Q, r_{<t}),
  \end{equation}
\noindent where $r_t$ represents the $t$-th token in the reasoning trace $R_{agg}$ and $\theta$ are the model parameters. The model learns to generate structured reasoning following the format established during cross-modal aggregation: \texttt{<think>...</think>} \texttt{<answer>...</answer>}, where the thinking section contains the multimodal reasoning and the answer section provides the final response.

\noindent\textbf{Stage 2: Reinforcement Learning.}
In a second step, we employ Group Relative Policy Optimization (GRPO)\cite{shao2024deepseekmath}. GRPO eliminates the need for explicit value function estimation by deriving advantage estimates through group-based comparisons of model outputs.

The GRPO training operates by sampling $G$ distinct responses $\{o_1, o_2, \ldots, o_G\}$ for each input question $q$ using the current policy $\pi_{\theta_{old}}$. Each response $o_i$ receives a scalar reward $r_i$ from our reward function. The advantage for $o_i$ is computed by normalizing rewards within the group:
\begin{equation}
    \hat{A}_{i, t} = \widetilde{r}_i = \frac{r_i - \text{mean}(\mathbf{r})}{\text{std}(\mathbf{r})},
\end{equation}
where this advantage $\widetilde{r}_i$ is applied uniformly across all tokens $t$ in response $o_i$.

Our reward combines multiple components to enforce correctness, formatting, and reasoning quality:
\begin{equation}
r_i = R_{format}(o_i) + R_{acc}(o_i) + R_{length}(o_i) .
\end{equation}

\noindent The three components are defined as:

\noindent\textbf{(1) Format Reward ($R_{format}$):} A binary reward that verifies adherence to our proposed reasoning format (\texttt{<think>...</think><answer>...</answer>}):
\begin{equation}
    R_{format}(o_i) = \begin{cases}
    1, & \text{if format is correct} \\
    0, & \text{otherwise}
    \end{cases}
\end{equation}
\textbf{(2) Final Answer Accuracy ($R_{acc}$):} A simple string matching evaluation that compares the model's predicted answer choice against the ground truth label:

\begin{equation}
R_{acc}(o_i) = \begin{cases}
1, & \text{if answer is correct}, \\
0, & \text{otherwise}.
\end{cases}
\end{equation}

\noindent\textbf{(3) Reasoning Length Reward ($R_{length}$):} A dense reward that encourages optimal reasoning trace length using a Gaussian-shaped distribution. The reward is:
\begin{equation}
\begin{aligned}
R_{\text{length}}(o_i)
= \min\Big(
    1.0,\;
    &\exp\Big(-\frac{(w_i - \mu)^2}{2\sigma^2}\Big) \\
    &\quad + \mathbb{I}(w_{\min} \le w_i \le w_{\max}) \cdot b
\Big)
\end{aligned}
\end{equation}
where $w_i$ is word count, $\mu$ is the optimal target, $\sigma$ is the width, and $b$ is a bonus in $[w_{min}, w_{max}]$.

We optimize our reward function using Group Relative Policy Optimization (GRPO) \cite{shao2024deepseekmath}, which computes advantage estimates through group-based comparisons of model outputs and applies policy gradient optimization with clipping and KL regularization.

\section{Experiments}

\subsection{Training Datasets}
\textbf{AVRT-20K for Supervised fine-tuning.}
In the first phase of the training, we fine-tune the respective student model in a fully supervised way on pre-constructed audio-visual reasoning traces. To this end, we introduce the AVRT-20K dataset, which is constructed using our proposed AVRT method on a subset of the AVQA dataset. We use Kimi-VL-Thinking \cite{kimiteam2025kimivltechnicalreport} and Audio Flamingo 3  (\textit{think}) \cite{goel2025audio} as the single-modality teachers $T_V$ and $T_A$. These models were chosen due to their balance between achieving state-of-the-art results in their modalities, and generating descriptive reasoning traces. We use 10-second audio input and 8 uniformly-sampled video frames from each sample as the input for the audio and visual teacher, respectively. The full prompt templates used for each model can be found in the supplementary material.

The final collection comprises $18,279$ training samples and $945$ validation samples, all extracted from the original AVQA dataset.
All samples achieve $100\%$ reasoning format compliance, ensuring consistent structure across the dataset. The thinking sections contain an average of $165.5 \pm 33.9$ tokens in the training set and $163.4 \pm 32.5$ tokens in the validation set, while answer sections are consistently single tokens, corresponding to the (A, B, C, D) options format. Videos maintain uniform duration of $\approx$10 seconds, with the main resolution being 1280$\times$720.
Table~\ref{tab:dataset_stats} presents statistics for our AVRT-20K dataset.

\begin{table}[t]
    \centering
    \footnotesize
    \setlength{\tabcolsep}{5pt}
    \centering
    \begin{tabular}{@{}l r@{}}
    \toprule
    \textbf{AVRT-20K Metrics} & \textbf{Train / Val} \\
    \midrule
    Total Samples & 18{,}279 / 945 \\
    Reasoning Format Compliance & 100.0\% / 100.0\% \\
    Thinking Section Length (tokens) & $165.5 \pm 33.9$ \,/\, $163.4 \pm 32.5$ \\
    Answer Section Length (tokens) & $1.0 \pm 0.0$ \,/\, $1.0 \pm 0.0$ \\
    Video and Audio Duration (sec) & $10.0 \pm 0.1$ \,/\, $10.0 \pm 0.2$ \\
    Primary Resolution & 1280$\times$720 (62\% \,/\, 43\%) \\
    \bottomrule
    \end{tabular}
    
    \caption{Statistics of the AVRT-20K dataset showing sample counts, formatting, token length as well as data properties for the training and validation split.}
    \label{tab:dataset_stats}
    
\end{table}

The distribution of resulting question types closely mirrors that of the original AVQA dataset, with "Which" questions being most prevalent ($45.2\%$ in training), followed by "Come From" ($30.9\%$), "Happening" ($15.5\%$), and "Where" ($8.0\%$) questions showing that the sampling successfully encompasses the distributional characteristics of the original dataset, ensuring our subset maintains representativeness across different reasoning types and question categories.

\noindent\textbf{RL Training.}
In the second phase, the resulting model is trained on the full AVQA training set \cite{yang2022avqa}. AVQA comprises $57,335$ question-answer pairs across $45,867$ unique audio-visual samples from daily activities. The dataset is split into $40,127$ training samples and $17,208$ validation samples. For the RL phase, we use the full training set to further improve the model's audio-visual reasoning capabilities through reinforcement learning, building upon the reasoning foundations established during the SFT phase.

\subsection{Benchmark Datasets}
We evaluate the resulting models on a diverse set of benchmarks spanning different modality combinations to assess cross-modal reasoning capabilities.

\noindent\textbf{Audiovisual.}
\textit{AVQA} \cite{yang2022avqa} is a large-scale benchmark containing $57{,}015$ question-answer pairs across $45{,}867$ videos designed to evaluate models' ability to reason over both audio and visual content. We note that AVQA serves as an \textit{in-domain} evaluation, since our training data is derived from its training set. \\
\textit{DailyOmni} \cite{zhou2025dailyomniaudiovisualreasoningtemporal} evaluates models on real-life audio-visual scenarios that require joint reasoning across video, audio, and textual information. The dataset contains $684$ videos and $1{,}197$ question-answer pairs ($550$ from $60$-second videos, $647$ from $30$-second videos) covering all $11$ YouTube categories.\\
\textit{OmniBench} \cite{li2024omnibench} evaluates models on integrate image, audio, and text inputs for cross-modal reasoning. The benchmark contains $1{,}142$ question-answer pairs organized into $8$ categories, with audio clips averaging $9.22$ seconds in duration.\\
\textit{Video-MME} \cite{fu2025video} is a comprehensive multi-modal evaluation benchmark for video analysis, spanning $6$ primary visual domains with $30$ subfields, covering short-, medium-, and long-term videos ranging from $11$ seconds to $1$ hour, with $2{,}700$ question-answer pairs across $900$ videos. While commonly adopted as a general video understanding benchmark, Video-MME also integrates audio information, making it suitable for evaluating audio-visual capabilities. In our evaluation, all models receive both the video frames and the audio track as input.\\
\textit{RivaBench} \cite{sun2025video} extends the scope of complex video understanding with reasoning-intensive application scenarios, including academic presentations (Academic) and stand-up comedy (StandUp). It contains $1{,}912$ and $2{,}128$ five-way multiple-choice questions for the Academic and StandUp scenarios, respectively, with mean video durations of $47.2$s and $43.2$s sourced from YouTube.

\noindent\textbf{Audio-Only.}
\textit{MMAR} \cite{ma2025mmar} is an audio reasoning benchmark designed to evaluate models' ability to perform complex reasoning tasks on auditory information.\\
\textit{MMAU} \cite{sakshi2024mmau} evaluates multimodal audio understanding on tasks requiring expert-level knowledge and complex reasoning. It comprises $10$k audio clips paired with human-annotated questions spanning speech, environmental sounds, and music, covering information extraction and reasoning tasks across $27$ distinct skills.\\
We include both audio benchmarks to assess how well the proposed cross-modal training approach transfers to single-modality audio reasoning scenarios.
\subsection{Implementation Details}
For all experiments, we use Qwen2.5-Omni \cite{Qwen2.5-Omni} as the base student models with frozen vision and audio modules. Supervised fine-tuning is conducted on $18,279$ samples over $1$ epoch with an effective batch size of $32$ (1 sample per device $\times$ $8$ gradient accumulation steps $\times$ $4$ H100 GPUs). We use a learning rate of $2e-6$ with cosine scheduling, AdamW optimizer ($\beta_1=0.9, \beta_2=0.999, \epsilon=1e-8$), weight decay of $0.01$, and $100$ warmup steps. Training employs DeepSpeed ZeRO Stage 2 optimization with CPU offloading and bfloat16 precision. 
For reinforcement learning, we use identical infrastructure with GRPO-specific hyperparameters: group size $G=4$, clipping parameter $\epsilon=0.2$, KL regularization coefficient $\beta=0.01$, and temperature 1. For the reasoning length reward, the optimal target length is set to $\mu = 100$ words with a standard deviation of $\sigma = 20$ words, and the target range for the bonus to $w_{min} = 100$ and $w_{max} = 200$ words.
\subsection{Comparison to State-of-the-Art}
Table~\ref{tab:main_results} shows the performance of the proposed AVRT-trained model across seven benchmark datasets spanning different modality combinations. We include Qwen3-Omni \cite{xu2025qwen3} (30B parameters) as an upper bound to contextualize improvements, and report the single-modality teachers as reference points for our distillation process. Our 3B- and 7B-parameter models achieve strong performance both in absolute accuracy and relative improvement over base models, with our results frequently surpassing the individual teacher models despite using only single-modality specialists during training.
\begin{table*}[t]
\centering
\resizebox{\textwidth}{!}{%
\begin{tabular}{lc|c||cccccc|ccc}
\toprule
\multirow[c]{2}{*}{\textbf{Model}} & \multicolumn{1}{c}{\multirow[c]{2}{*}{\includegraphics[height=15pt]{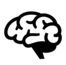}}} & \multicolumn{7}{c}{\textit{Audio-Visual}} & \multicolumn{3}{c}{\textit{Audio}} \\
\cmidrule(lr){3-9} \cmidrule(lr){10-12}
& \multicolumn{1}{c}{} & \multicolumn{1}{c}{\textbf{AVQA}$^\dagger$} & \textbf{DailyOmni} & \textbf{OmniBench} & \textbf{Video-MME} & \textbf{Riva-S} & \textbf{Riva-A} & \multicolumn{1}{c}{\textit{AV Avg.}} & \textbf{MMAR} & \textbf{MMAU} & \multicolumn{1}{c}{\textit{A Avg.}} \\
\hline
\rowcolor{gray!20} \multicolumn{12}{c}{\textit{30B Audio-Visual Models}} \\
\textcolor{gray!50}{Qwen3-Omni$^*$ \cite{xu2025qwen3}} & \textcolor{gray!50}{\checkmark} & \textcolor{gray!50}{93.8} & \textcolor{gray!50}{57.5} & \textcolor{gray!50}{63.6} & \textcolor{gray!50}{57.5} & \textcolor{gray!50}{81.7} & \textcolor{gray!50}{56.2} & \textcolor{gray!50}{63.3} & \textcolor{gray!50}{66.4} & \textcolor{gray!50}{76.5} & \textcolor{gray!50}{71.5} \\
\hline
\rowcolor{gray!20} \multicolumn{12}{c}{\textit{Modality-Specific Teachers}} \\
\textcolor{gray!50}{Kimi-VL-Thinking$^*$ \cite{kimiteam2025kimivltechnicalreport}} & \textcolor{gray!50}{\checkmark} & \textcolor{gray!50}{81.4} & \textcolor{gray!50}{47.0} & \textcolor{gray!50}{33.5} & \textcolor{gray!50}{58.9} & \textcolor{gray!50}{72.3} & \textcolor{gray!50}{49.6} & \textcolor{gray!50}{52.3} & \textcolor{gray!50}{\textit{N/A}} & \textcolor{gray!50}{\textit{N/A}} & \textcolor{gray!50}{\textit{N/A}} \\
\textcolor{gray!50}{AF3 \textit{(think)$^*$} \cite{goel2025audio}} & \textcolor{gray!50}{\checkmark} & \textcolor{gray!50}{64.3} & \textcolor{gray!50}{52.5} & \textcolor{gray!50}{28.9} & \textcolor{gray!50}{45.0} & \textcolor{gray!50}{65.5} & \textcolor{gray!50}{41.7} & \textcolor{gray!50}{46.7} & \textcolor{gray!50}{60.1} & \textcolor{gray!50}{73.3} & \textcolor{gray!50}{66.7} \\

\hline
\rowcolor{gray!20} \multicolumn{12}{c}{\textit{3B Audio-Visual Models}} \\
AVATAR \cite{kulkarni2025avatarreinforcementlearningsee} & \checkmark & - & \underline{44.7} & 45.8 & - & - & - & - & - & - & - \\
Qwen2.5 Omni$^*$ \cite{Qwen2.5-Omni} & × & \underline{88.3} & 43.1 & \underline{50.2} & \underline{55.4} & \underline{62.4} & \underline{38.4} & \underline{49.9} & \underline{53.7} & \underline{61.1} & \underline{57.4} \\
\rowcolor{almond}\textbf{AVRT (Ours)} & \checkmark & \textbf{91.1} \tiny{\textcolor{darkgreen}{(+2.8)}} & \textbf{49.2} \tiny{\textcolor{darkgreen}{(+6.1)}} & \textbf{56.3} \tiny{\textcolor{darkgreen}{(+6.1)}} & \textbf{62.6} \tiny{\textcolor{darkgreen}{(+7.2)}} & \textbf{71.3} \tiny{\textcolor{darkgreen}{(+8.9)}} & \textbf{49.3} \tiny{\textcolor{darkgreen}{(+10.9)}} & \textbf{57.7} \tiny{\textcolor{darkgreen}{(+7.8)}} & \textbf{57.3} \tiny{\textcolor{darkgreen}{(+3.6)}} & \textbf{70.0} \tiny{\textcolor{darkgreen}{(+8.9)}} & \textbf{63.7} \tiny{\textcolor{darkgreen}{(+6.3)}} \\
\hline
\rowcolor{gray!20} \multicolumn{12}{c}{\textit{7B Audio-Visual Models}} \\
EchoInk \cite{xing2025echoinkr1exploringaudiovisualreasoning} & \checkmark & - & 46.2 & 46.5 & - & - & - & - & - & - & - \\
Omni-R1 \cite{zhong2025omnir1reinforcementlearningomnimodal} & \checkmark & - & 46.8 & 46.9 & 60.7 & - & - & - & - & - & - \\
HumanOmni \cite{zhao2025humanomni} & \checkmark & - & 47.6 & 44.9 & - & - & - & - & - & - & - \\
Ola-7B \cite{liu2025ola} & × & - & 52.3 & 45.3 & \textbf{68.4} & - & - & - & - & - & - \\
AV-Reasoner \cite{lu2025avreasonerimprovingbenchmarkingcluegrounded} & \checkmark & - & 53.8 & 48.3 & - & - & - & - & - & - & - \\
AVATAR \cite{kulkarni2025avatarreinforcementlearningsee} & \checkmark & - & 47.0 & 49.1 & - & - & - & - & - & - & - \\
V-RTS$^*$ \cite{wang2025video} & \checkmark & \underline{86.6} & 47.8 & 35.3 & 63.0 & 70.3 & \underline{48.9} & 53.1 & \textit{N/A} & \textit{N/A} & \textit{N/A} \\
v-SALMONN-o1$^*$ \cite{sun2025video} & \checkmark & 84.8 & \textbf{64.0} & 40.5 & 41.3 & \textbf{76.7} & 48.3 & 54.2 & 51.2 & 53.8 & 52.5 \\
Qwen2.5 Omni$^*$ \cite{Qwen2.5-Omni} & × & 84.9 & 51.5 & \underline{50.7} & 58.1 & 74.9 & 41.3 & 55.3 & \underline{56.5} & \underline{71.9} & 64.2 \\
\rowcolor{almond}\textbf{AVRT (Ours)} & \checkmark & \textbf{90.4} \tiny{\textcolor{darkgreen}{(+5.5)}} & \underline{54.4} \tiny{\textcolor{darkgreen}{(+2.9)}} & \textbf{57.1} \tiny{\textcolor{darkgreen}{(+6.4)}} & \underline{64.1} \tiny{\textcolor{darkgreen}{(+6.0)}} & \underline{75.3} \tiny{\textcolor{darkgreen}{(+0.4)}} & \textbf{50.7} \tiny{\textcolor{darkgreen}{(+9.4)}} & \textbf{60.3} \tiny{\textcolor{darkgreen}{(+5.0)}} & \textbf{59.1} \tiny{\textcolor{darkgreen}{(+2.6)}} & \textbf{75.4} \tiny{\textcolor{darkgreen}{(+3.5)}} & \textbf{67.3} \tiny{\textcolor{darkgreen}{(+3.1)}} \\
\bottomrule
\end{tabular}%
}

\caption{Comparison of audio-visual reasoning models on benchmark datasets. AVQA is considered in-domain since the training dataset is derived from its training set, and therefore excluded from the average. All other models are tested in zero-shot mode without finetuning. Reproduced baseline results marked with $^*$. \includegraphics[height=9pt]{figs/brain_v2.png} stands for reasoning.}
\label{tab:main_results}

\end{table*}

\noindent\textbf{Audio-Visual Benchmarks.}
On all datasets, the 3B model significantly outperforms the base model, surpassing even some 7B models, e.g., on DailyOmni with $49.2\%$ {\scriptsize\textcolor{darkgreen}{(+6.1)}} (improvements w.r.t.\ Qwen2.5 Omni baselines) and OmniBench with $56.3\%$ {\scriptsize\textcolor{darkgreen}{(+6.1)}} accuracy respectively. %
The 7B variant achieves $54.4\%$ {\scriptsize\textcolor{darkgreen}{(+2.9)}} on DailyOmni and $57.1\%$ {\scriptsize\textcolor{darkgreen}{(+6.4)}} on OmniBench, outperforming most baselines with the exception of video-SALMONN-o1, which achieves $64.0\%$ on DailyOmni. This gap reflects a difference in specialization as video-SALMONN-o1 is trained on speech-heavy data, which aligns well with DailyOmni's dialogue and narration tasks, whereas AVRT targets general sound reasoning. On benchmarks beyond speech, AVRT consistently outperforms video-SALMONN-o1 as for OmniBench (57.1 vs.\ 40.5), MMAR (59.1 vs.\ 51.2), and MMAU (75.4 vs.\ 53.8).
For RivaBench, the 3B model achieves $71.3\%$ {\scriptsize\textcolor{darkgreen}{(+8.9)}} on StandUp (Riva-S) and $49.3\%$ {\scriptsize\textcolor{darkgreen}{(+10.9)}} on Academic (Riva-A). The 7B variant achieves $75.3\%$ {\scriptsize\textcolor{darkgreen}{(+0.4)}} on StandUp and $50.7\%$ {\scriptsize\textcolor{darkgreen}{(+9.4)}} on Academic, showing that performance improvements scale across model sizes. The large gains on Academic demonstrate that the reasoning approach is particularly effective on expert-level content requiring deeper understanding of technical material.
For AVQA, an in-domain evaluation since our training data is derived from its training set, the 3B model achieves $91.1\%$ {\scriptsize\textcolor{darkgreen}{(+2.8)}} accuracy, while the 7B variant reaches $90.4\%$ {\scriptsize\textcolor{darkgreen}{(+5.5)}}. The results validate that reasoning-based supervision provides value beyond simple data augmentation.

\noindent\textbf{Audio Benchmarks.}
On audio-only reasoning, the model shows strong transfer capabilities. For MMAR, the 3B model achieves $57.3\%$ {\scriptsize\textcolor{darkgreen}{(+3.6)}} and approaches the specialized audio teacher AF3 \textit{(think)} ($60.1\%$), while the 7B model reaches $59.1\%$ {\scriptsize\textcolor{darkgreen}{(+2.6)}}. On MMAU, AVRT 3B achieves $70.0\%$ {\scriptsize\textcolor{darkgreen}{(+8.9)}} and the 7B model achieves $75.4\%$ {\scriptsize\textcolor{darkgreen}{(+3.5)}}, demonstrating that audio reasoning capabilities learned through multimodal training transfer effectively to complex single-modality tasks.
Overall, AVRT 3B achieves an audio-visual average of $57.7\%$ {\scriptsize\textcolor{darkgreen}{(+7.8)}} and an audio average of $63.7\%$ {\scriptsize\textcolor{darkgreen}{(+6.3)}}, while AVRT 7B reaches $60.3\%$ {\scriptsize\textcolor{darkgreen}{(+5.0)}} and $67.3\%$ {\scriptsize\textcolor{darkgreen}{(+3.1)}} respectively. These consistent improvements across diverse benchmarks, dataset sizes, and modality combinations validate the importance of structured reasoning in multimodal tasks and demonstrate that effective audio-visual reasoning can be achieved through single-modality teacher composition.
We note that the 7B model shows smaller relative gains than the 3B model (e.g., AV Avg.\ +5.0 vs.\ +7.8), which is expected since the 7B baseline is already stronger (e.g., DailyOmni 51.5 vs.\ 43.1), leaving less headroom for improvement. This pattern is consistent with distillation literature, where smaller or weaker models benefit more from teacher guidance \cite{busbridge25distillation}. Importantly, the 7B model still improves across all benchmarks, confirming that the approach is not limited to small model sizes.

\subsection{Ablation Studies}
\label{sec:ablation}

\noindent\textbf{Evaluation of SFT fine-tuning.} We first assess the impact of the supervised fine-tuning step on the generated reasoning traces compared to the Qwen2.5-Omni 3B baseline, as well as to the same model trained only with an RL objective. As shown in Table~\ref{tab:ablation_single_modality}, simply training the model with an RL objective leads to an improvement of $2.1\%$ (52.3\% vs. 50.2\%) on the audio-visual setting. The proposed 2-stage training with a SFT cold-start based on the generated reasoning traces further improves performance to $56.3\%$, an additional gain of $4.0$ points over the RL-only baseline. We hypothesize that SFT provides two key benefits for the subsequent RL stage. First, \textit{format learning}: SFT teaches the model the structured reasoning format (thinking section followed by answer), ensuring high $R_{format}$ and $R_{length}$ rewards from the start of RL training. Second, \textit{reasoning priors}: The model learns cross-modal correlation patterns from the distilled traces, which guide RL exploration toward productive reasoning strategies rather than exploring randomly.

\noindent\textbf{Single Modality Performance.}
To further investigate the impact of training on multimodal data on the respective single-modality performance, we evaluate the model using one modality at a time.
As shown in Table~\ref{tab:ablation_single_modality}, the results demonstrate that multimodal training with the proposed reasoning trace aggregation approach not only leads to improvements on audio-visual settings, but also on the single-modality performance.
This can be considered as an indication for reasoning transfer learning as the SFT dataset is composed mainly of questions that require both audio and vision ($99.0\%$) (see supplementary material) and as both our supervised fine-tuning (SFT) with reasoning traces plus RL and the RL-only baseline are trained solely on audio-visual inputs, without any modality dropout or single-modality augmentation.
\begin{table*}[!t]
\centering

\begin{minipage}{0.45\textwidth}
\centering
\resizebox{\textwidth}{!}{%
\begin{tabular}{ccccc}
\toprule
\textbf{Model} & SFT & RL & \textbf{Mod.} & \textbf{OmniBench} \\
\midrule
\multirow{3}{*}{\begin{tabular}{@{}c@{}}Qwen2.5-Omni 3B\\(Baseline)\end{tabular}} & × & × & A & 39.4 \\
& × & × & V & 42.7 \\
& × & × & AV & 50.2 \\
\midrule
\multirow{3}{*}{\begin{tabular}{@{}c@{}}Qwen2.5-Omni 3B + RL\\(Baseline+RL)\end{tabular}} & × & \checkmark & A & 41.1 \scriptsize{\textcolor{darkgreen}{(+1.7)}} \\
& × & \checkmark & V & 43.2 \scriptsize{\textcolor{darkgreen}{(+0.5)}} \\
& × & \checkmark & AV & 52.3 \scriptsize{\textcolor{darkgreen}{(+2.1)}} \\
\midrule
\rowcolor{almond} & \checkmark & \checkmark & A & \textbf{41.9} \scriptsize{\textcolor{darkgreen}{(+2.5)}} \\
\rowcolor{almond}\textbf{AVRT (Ours)} & \checkmark & \checkmark & V & \textbf{45.8} \scriptsize{\textcolor{darkgreen}{(+3.1)}} \\
\rowcolor{almond} & \checkmark & \checkmark & AV & \textbf{56.3} \scriptsize{\textcolor{darkgreen}{(+6.1)}} \\
\bottomrule
\end{tabular}%
}

\caption{Ablation of training stages on OmniBench for only audio(A), only vision(V), and audio-visual input(AV).}
\label{tab:ablation_single_modality}

\end{minipage}%
\hfill
\begin{minipage}{0.51\textwidth}

\centering
\resizebox{\textwidth}{!}{%
\begin{tabular}{lc}
\toprule
\textbf{Reasoning Chain Modality} & \textbf{OmniBench} \\
\midrule
Qw2.5-O-3B + RL & 52.3 \\
Qw2.5-O-3B + SFT Audio-Only + RL & 51.0 \scriptsize{\textcolor{red}{(-1.3)}} \\
Qw2.5-O-3B + SFT Video-Only + RL & 52.1 \scriptsize{\textcolor{darkgreen}{(-0.2)}} \\
\rowcolor{almond}\textbf{Audio-Visual (Ours)} & \textbf{56.3} \scriptsize{\textcolor{darkgreen}{(+4.0)}} \\
\bottomrule
\end{tabular}%
}

\caption{Ablation on reasoning chain types for SFT. Models are fine-tuned on audio-only resp. video-only reasoning chains, then trained with AV-RL. Our audio-visual aggregation outperforms single-modality chains.}
\label{tab:ablation_reasoning_chains}

\end{minipage}
\begin{minipage}{0.45\textwidth}
\centering
\resizebox{\textwidth}{!}{%
\begin{tabular}{lcc}
\toprule
\textbf{LLM Merger } & \textbf{Student Model} & \textbf{OmniBench} \\
\midrule
Gemma3-12B-It & Qwen2.5-Omni-3B & 48.5 \\
\rowcolor{almond}Qwen2.5-14B-Instruct & Qwen2.5-Omni-3B & \textbf{56.3} \\
\bottomrule
\end{tabular}%
}

\caption{Ablation on merger models. An LLM merger from the same model family as the student enhances performance.}
\label{tab:ablation_mergers}

\end{minipage}%
\hfill
\begin{minipage}{0.51\textwidth}
\centering
\resizebox{\textwidth}{!}{%
\begin{tabular}{lcc}
\toprule
\textbf{Reward Components} & \textbf{DailyOmni} & \textbf{OmniBench} \\
\midrule
Baseline & 43.1 & 50.2 \\
$R_{acc} + R_{format}$ & 45.5 \scriptsize{\textcolor{darkgreen}{(+2.4)}} & 54.7 \scriptsize{\textcolor{darkgreen}{(+4.5)}} \\
\rowcolor{almond}$R_{acc} + R_{format} + R_{length}$ & \textbf{49.2} \scriptsize{\textcolor{darkgreen}{(+6.1)}} & \textbf{56.3} \scriptsize{\textcolor{darkgreen}{(+6.1)}} \\
\bottomrule
\end{tabular}%
}

\caption{Ablation on reasoning chain length reward. Including a length reward consistently improves performance.}
\label{tab:ablation_reward_length}

\end{minipage}

\end{table*}

\noindent\textbf{Reasoning trace types.} To validate the effectiveness of the audio-visual reasoning trace aggregation, we compare the approach against alternative supervision strategies. As shown in Table~\ref{tab:ablation_reasoning_chains}, we evaluate models trained with: (1) no SFT (RL only, +2.1), (2) SFT on audio-only reasoning traces (+0.8), (3) SFT on video-only reasoning traces (+1.9), and (4) audio-visual aggregated traces (+6.1), all compared to the Qwen2.5-Omni 3B baseline (50.2\%). The audio-visual reasoning traces achieve the best performance at 56.3\%, substantially outperforming single-modality alternatives. This demonstrates that aggregating diverse reasoning perspectives from both modalities provides more effective supervision than training on single-modality traces, which may introduce modality-specific biases that hinder cross-modal integration.

\noindent\textbf{Different merger models.} We investigate the impact of using different teacher models for cross-modal aggregation in our pipeline. As shown in Table~\ref{tab:ablation_mergers}, we compare two merger models: Gemma3-12B-It and Qwen2.5-14B-Instruct. While Qwen2.5-14B-Instruct improves over the Qwen2.5-Omni 3B baseline (50.2\%) to achieve 56.3\% (+6.1 points), Gemma3-12B-It decreases performance to 48.5\% (-1.7 points). The 7.8-point gap reveals the importance of architectural alignment. During training, Qwen2.5-14B-Instruct converged significantly faster to the MCQ format than Gemma3, which we attribute to tokenization alignment. Recent work on reasoning transfer~\cite{bousselham2025vold} shows that shared architecture between teacher and student is critical for effective distillation. Here, using a merger with the same backbone as the student enables aligned token distributions and faster convergence, whereas the Gemma3 merger's different token patterns force the student to adapt to a substantially different distribution during fine-tuning.

\noindent\textbf{Reasoning length reward.} Next, we consider the impact of incorporating a reward function that encourages optimal reasoning trace length. As shown in Table~\ref{tab:ablation_reward_length}, compared to the Qwen2.5-Omni 3B baseline (DailyOmni 43.1\%, OmniBench 50.2\%), the format and accuracy rewards alone achieve 45.5\% and 54.7\% respectively. Adding the length reward further improves to 49.2\% (+6.1 points on DailyOmni) and 56.3\% (+6.1 points on OmniBench). This consistent improvement demonstrates that maintaining a sufficient reasoning trace length is effective for audio-visual reasoning. Notably, the Gaussian-shaped reward penalizes both too-short and too-long responses, discouraging runaway verbosity rather than monotonically rewarding length; the moderate target of $\mu{=}100$ words encourages concise but adequate reasoning depth.

\noindent\textbf{Dual-Teacher Filtering Strategy.} Our approach filters training data to include only samples where both single-modality teachers produce correct answers, reducing the dataset from 40K to 20K samples (50\% filtered). A potential concern is that this filtering might introduce bias toward ``easy'' samples. To validate that filtering does not limit generalization, we compare filtered vs.\ unfiltered training on OmniBench. Training with unfiltered data (40K samples, matched batch sizes) yields $53.6\%$ accuracy, while our filtered approach achieves $56.3\%$ ($-2.7$ points for unfiltered). This demonstrates that the dual-teacher filtering strategy improves performance by selecting high-quality reasoning traces, and the hard questions in OmniBench (54\% of test set) confirm the model generalizes beyond the training distribution despite filtering. We also note that question types where both teachers systematically fail would be absent from SFT supervision. However, since the two teachers operate on different modalities, their failure modes are largely decorrelated: the audio teacher tends to fail on visually-dependent questions while the visual teacher fails on audio-dependent ones, limiting systematic shared blind spots.

\begin{wraptable}{r}{4.2cm}

\centering
\begin{tabular}{lccc}
\toprule
\textbf{Model} & \textbf{Easy} & \textbf{Med} & \textbf{Hard} \\
\midrule
Baseline & 70.5 & 53.5 & 45.3 \\
\rowcolor{almond}\textbf{AVRT} & \textbf{76.9} & \textbf{59.2} & \textbf{51.8} \\
\bottomrule
\end{tabular}

\caption{Performance on different OmniBench difficulty subsets.}
\label{tab:omnibench_subsets}

\begin{tabular}{lcc}
\toprule
\textbf{Model} & \textbf{IFE} & \textbf{Logic Err} \\
& \footnotesize{\textbf{(\%) $\downarrow$}} & \footnotesize{\textbf{(\%) $\downarrow$}} \\
\midrule
Baseline & 0.9 & 48.9 \\
\rowcolor{almond}\textbf{AVRT} & \textbf{0.2} & \textbf{43.0} \\
\bottomrule
\end{tabular}

\caption{Error analysis on OmniBench: Instruction Format Error (IFE) and Logic Error rates.}
\label{tab:format_learning_analysis}

\end{wraptable}

\noindent\textbf{Omnibench Subsets.} To further validate that the dual-teacher filtering does not bias the model toward easy samples, we categorize OmniBench questions into subsets based on teacher-model performance: \textit{easy} (both teachers correct, 64 questions), \textit{medium} (one teacher correct, 456 questions), and \textit{hard} (neither correct, 619 questions). As shown in Table~\ref{tab:omnibench_subsets}, the model outperforms the Qwen2.5-Omni baseline across all difficulty levels: easy (76.9\% vs. 70.5\%, +6.4 points), medium (59.2\% vs. 53.5\%, +5.7 points), and hard (51.8\% vs. 45.3\%, +6.5 points). Crucially, the +6.5 point gain on hard questions, where neither teacher was correct, demonstrates that the student generalizes beyond the filtered training distribution. The \textit{easy} subset shows the largest improvement but exhibits greater variability due to its smaller sample size.

\noindent\textbf{Format Learning vs. Cross-Modal Reasoning.} %
To answer the question \textit{``Do improvements come from learning the output format or from reasoning?''}, we analyze error types on OmniBench. We classify each incorrect prediction into two categories: \textit{Instruction Format Error} (IFE), where the response does not follow the required answer format (\eg missing or malformed answer tags), and \textit{Logic Error}, where the response is correctly formatted but contains the wrong answer, indicating a reasoning failure. As shown in Table~\ref{tab:format_learning_analysis}, AVRT reduces logic errors by 5.9 points (48.9\% → 43.0\%) while reducing IFE by only 0.7 points. This demonstrates that performance gains come predominantly from enhanced cross-modal reasoning rather than format learning.

\subsection{Qualitative Results}
\begin{figure*}[t]
    \centering
    \includegraphics[width=\textwidth]{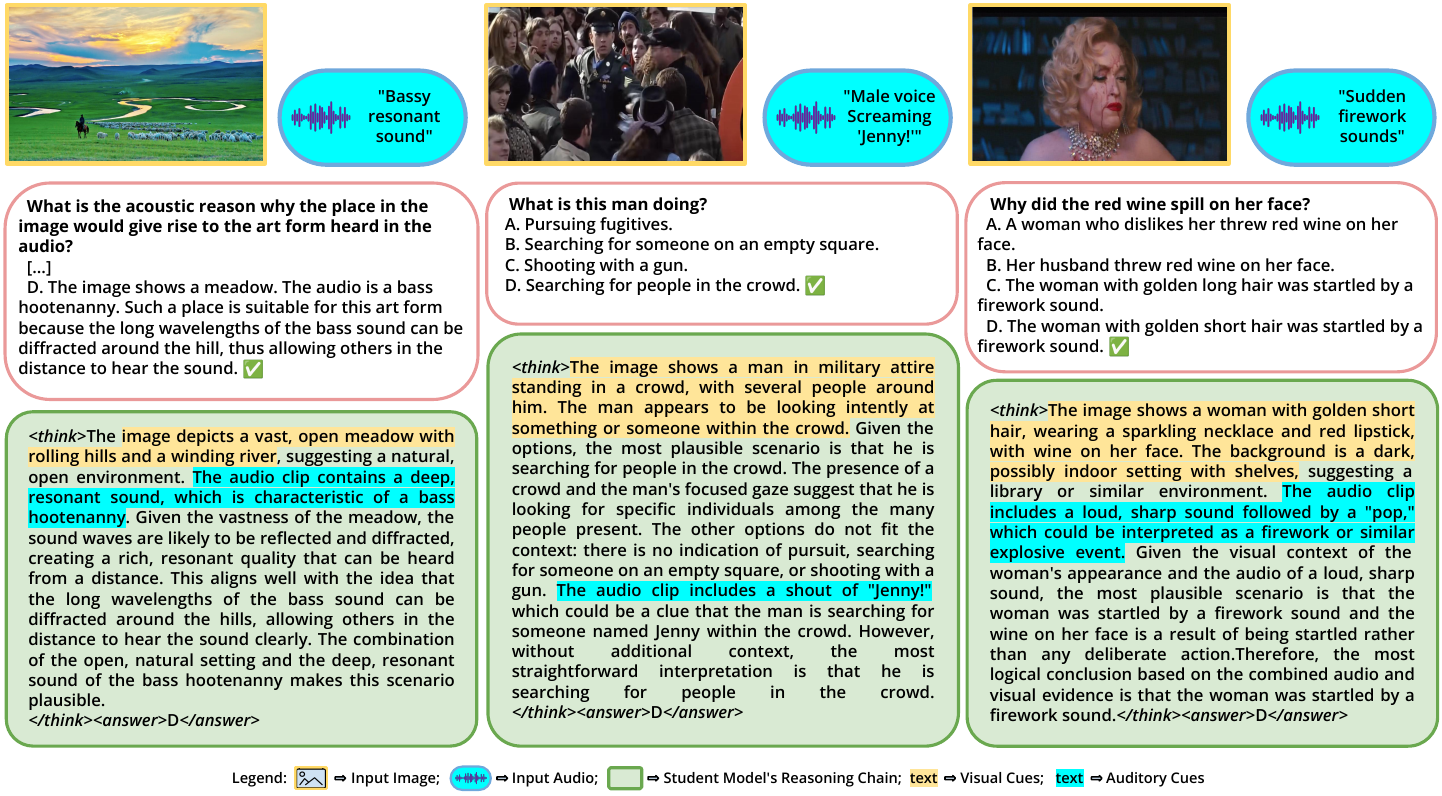}
    
    \caption{Qualitative results of the AVRT-trained model on OmniBench: It shows that the model trained on the respective AVRT-20K data is able to  retrieve audio and visual information to answer the question, to combine the two sources of information, and to generate high-quality reasoning chains based on different cues in both modalities. %
    }
    \label{fig:qualitative_results}
    
\end{figure*}

We finally provide qualitative results of reasoning traces produced by the student model after fine-tuning on the proposed AVRT reasoning traces on OmniBench in Fig.~\ref{fig:qualitative_results}.
The results show that with only SFT on the generated reasoning traces, the model is able to reason about the image and audio content to answer the question.
In the first example, the model correctly associates the acoustic properties of a bass ``hootenanny'' with the open meadow environment, demonstrating understanding of how sound propagates differently in open versus enclosed spaces.
The second example showcases more sophisticated multi-modal reasoning, where the model uses visual cues (people around the main character) to contextualize the audio (male voice calling ``Jenny!'') and correctly identifies the scenario as searching for people in a crowd among multiple plausible options.
The third example illustrates the model's ability to connect temporal audio events (firework sounds) with visual evidence (wine spill on face).
As shown in the figure, the model learns to incorporate both visual and auditory cues to arrive at correct answers. These results are on OmniBench, which is a particularly challenging dataset where $54\%$ of questions fall into the ``hard'' category (neither teacher correct). The model's reasoning traces demonstrate cross-modal integration rather than relying on single-modality shortcuts. Notably, the model successfully generalizes from its training domain (AVQA videos with 8 frames and audio) to a different evaluation domain (single static images paired with audio), suggesting that the reasoning capabilities learned through our teacher aggregation approach transfer effectively across different input formats and temporal structures.

\section{Conclusion}
We introduced AVRT, a framework that generates high-quality audio-visual reasoning traces by merging outputs from specialized single-modality teachers via a text-only LLM, and uses them as cold-start in a two-stage training pipeline.
Evaluated on seven benchmarks, the 3B model achieves state-of-the-art results among comparable-size models: 56.3\% on OmniBench, 49.2\% on DailyOmni, and 57.3\% on MMAR, with the 7B model further improving to 57.1\%, 54.4\%, and 59.1\%, respectively.
Beyond audio-visual tasks, cross-modal training also improves single-modality reasoning, providing evidence of genuine reasoning transfer.
While instantiated here for the audio-visual domain, the approach is applicable to any modality combination lacking reasoning data.

\bibliographystyle{splncs04}
\bibliography{main}

\appendix
\clearpage
\setcounter{page}{1}
\section*{Supplementary Material}
\numberwithin{figure}{section}
\numberwithin{table}{section}

This supplementary material is organized as follows. Section~\ref{app:additional_analyses} presents additional analyses that further investigate the source and nature of AVRT's improvements: Section~\ref{app:trace_quality} evaluates the quality and groundedness of the generated reasoning traces against a native multimodal reasoner (Qwen3-Omni-Thinking), including a hallucination source attribution analysis; Section~\ref{app:modality_consistency} examines model behavior under missing modalities; and Section~\ref{app:control_experiments} presents control experiments that systematically ablate the SFT data composition to isolate the contribution of reasoning trace content from output formatting and answer exposure. Section~\ref{app:additional_info} provides the complete dataset construction prompts and additional dataset statistics. Section~\ref{app:limitations_future} discusses limitations and future work.

\section{Additional Analyses} \label{app:additional_analyses}

\subsection{Reasoning Trace Quality Analysis} \label{app:trace_quality}

To assess the quality of our generated reasoning traces, we conduct an automated evaluation using Gemini-3.1-Flash-Lite as an external judge, comparing AVRT pipeline traces against those produced by Qwen3-Omni-Thinking~\cite{xu2025qwen3}, a native multimodal reasoning model. We perform two types of evaluation over ${\sim}600$ samples from the AVRT-20K validation set: (1)~\textit{individual evaluation}, where the judge independently classifies each model's trace into one of three categories: \textit{Accurate} (observations are factually correct and reasoning is sound), \textit{Missing Info} (correct answer but omits important evidence), or \textit{Hallucinating} (contains fabricated observations). The judge also reports per-modality grounding flags; and (2)~\textit{head-to-head comparison}, where the judge receives both traces simultaneously and selects a winner based on factual accuracy, completeness, logical soundness, and groundedness. The comparison prompt is shown in Figure~\ref{fig:quality_eval_prompt}.

\begin{figure}[t]
\centering
\includegraphics[width=\linewidth]{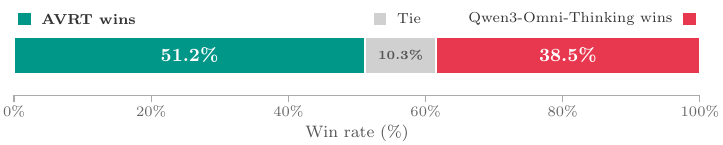}
\caption{\textbf{Head-to-head reasoning trace comparison on ${\sim}600$ AVRT-20K validation samples.}
Gemini-3.1-Flash-Lite judges AVRT traces as better in \textbf{51.2\%} of matchups vs.\ 38.5\% for Qwen3-Omni-Thinking, based on factual accuracy, completeness, and groundedness. The evaluation prompt is provided in Figure~\ref{fig:quality_eval_prompt}.}
\label{fig:head_to_head}
\end{figure}

Table~\ref{tab:trace_quality} and Figure~\ref{fig:head_to_head} summarize the results. In head-to-head comparison, AVRT traces are preferred in \textbf{51.2\%} of cases versus 38.5\% for Qwen3-Omni-Thinking (10.2\% ties). This is particularly notable because AVRT traces are generated by composing single-modality teachers, none of which has multimodal understanding on its own, while Qwen3-Omni-Thinking is a natively multimodal reasoning model with end-to-end audio-visual processing. In individual evaluation, both models achieve similar accuracy (${\sim}88\%$) and hallucination rates (${\sim}11\%$), confirming that AVRT's composed traces are on par with a native multimodal reasoning model. The most notable difference is in audio grounding: AVRT traces are grounded in audio content 93.4\% of the time, compared to 81.3\% for Qwen3-Omni-Thinking, suggesting that our explicit audio teacher stage produces more faithful acoustic descriptions. Qwen3-Omni-Thinking is slightly better at visual grounding (94.6\% vs.\ 93.4\%), but it exhibits a higher rate of \textit{Missing Info} (2.0\% vs.\ 0.3\%), indicating that it sometimes omits modality-specific evidence.

\begin{figure*}[t]
\centering
\small

\begin{minipage}[t]{0.48\textwidth}
\begin{tcolorbox}[colback=blue!5,colframe=blue!50!black,title={\scriptsize Audio Teacher: ``What are the people doing?'' \textbf{GT: A}},fonttitle=\scriptsize\bfseries,left=2pt,right=2pt,top=2pt,bottom=2pt]
\centering
\includegraphics[width=0.55\linewidth]{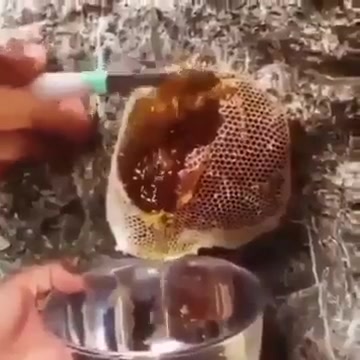}\\[0.3em]
\scriptsize\raggedright
\textbf{Pred: A (take honey)} \checkmark\\[0.2em]
\textbf{Audio:} ``The audio features the \underline{buzzing of bees} and \underline{honeycomb being handled}.''\\[0.2em]
\textbf{Vision:} Correctly identifies hands scraping honey from a honeycomb into a container.\\[0.2em]
\textbf{Merged:} ``The \textcolor{blue}{audio} reveals \underline{buzzing bees}\ldots the \underline{dripping sound} corresponds with the \textcolor{red}{visual} of honey flowing.''\\[0.2em]
\textit{\textcolor{gray}{Judge: No bee buzzing or dripping in the audio. Vision correct.}}
\end{tcolorbox}
\end{minipage}%
\hfill
\begin{minipage}[t]{0.48\textwidth}
\begin{tcolorbox}[colback=red!5,colframe=red!50!black,title={\scriptsize Vision Teacher: ``Source of sound?'' \textbf{GT: B (wind)}},fonttitle=\scriptsize\bfseries,left=2pt,right=2pt,top=2pt,bottom=2pt]
\centering
\includegraphics[width=0.55\linewidth]{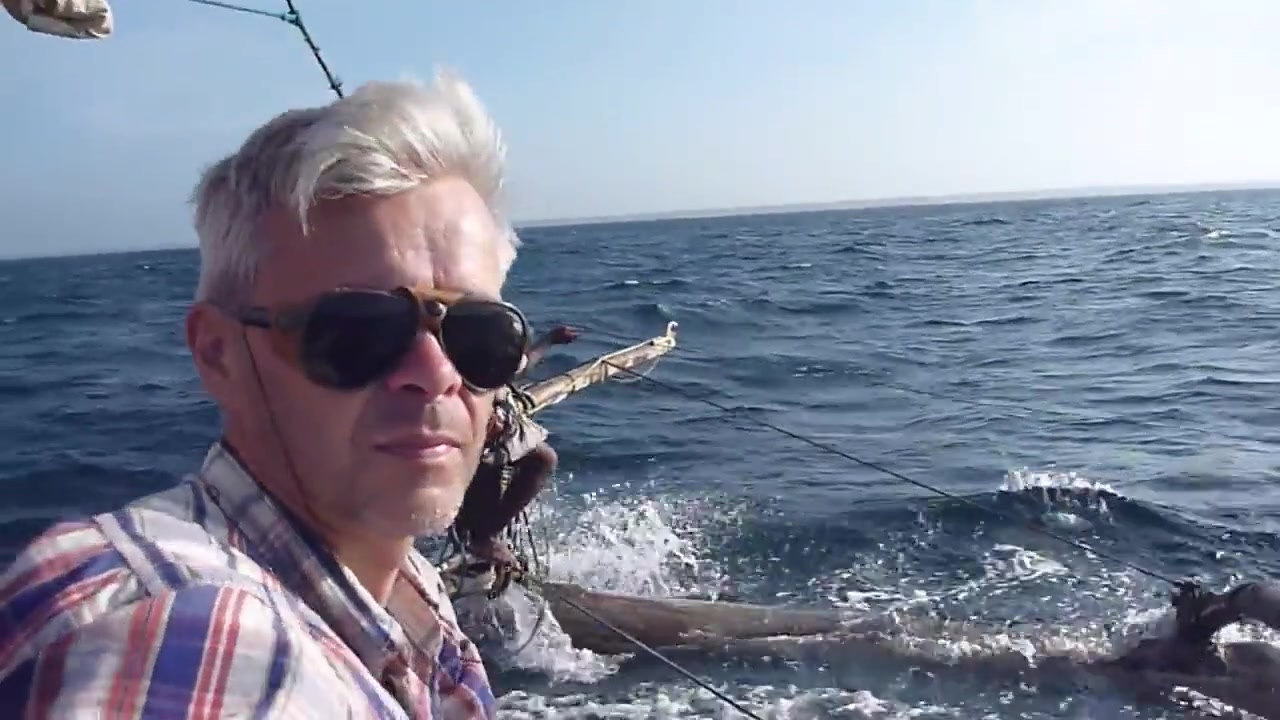}\\[0.3em]
\scriptsize\raggedright
\textbf{Pred: B} \checkmark\\[0.2em]
\textbf{Audio:} Correctly identifies the sound of wind.\\[0.2em]
\textbf{Vision:} ``A person navigating an outrigger canoe\ldots \underline{absence of sails}\ldots the individual \underline{using a paddle}.''\\[0.2em]
\textbf{Merged:} ``The individual \underline{using a paddle}\ldots the \underline{absence of a sail} makes it less likely to hear wind\ldots''\\[0.2em]
\textit{\textcolor{gray}{Judge: Video shows a sail and the boat under wind power. Vision misidentified the scene; merger incorporated false observations.}}
\end{tcolorbox}
\end{minipage}

\begin{minipage}[t]{0.48\textwidth}
\begin{tcolorbox}[colback=orange!5,colframe=orange!50!black,title={\scriptsize Both Teachers: ``What are the people doing?'' \textbf{GT: D}},fonttitle=\scriptsize\bfseries,left=2pt,right=2pt,top=2pt,bottom=2pt]
\centering
\includegraphics[width=0.55\linewidth]{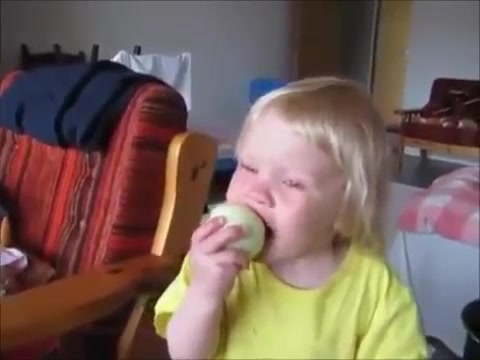}\\[0.3em]
\scriptsize\raggedright
\textbf{Pred: D (eat fruit)} \checkmark\\[0.2em]
\textbf{Audio:} ``The audio features the \underline{sound of someone eating fruit}.''\\[0.2em]
\textbf{Vision:} ``The child is holding a \underline{green, round object}\ldots consuming \underline{fruit}.''\\[0.2em]
\textbf{Merged:} ``\textcolor{blue}{Audio}: \underline{eating fruit}\ldots \textcolor{red}{Visual}: child holding a \underline{green object}\ldots eating fruit.''\\[0.2em]
\textit{\textcolor{gray}{Judge: Child is eating a raw onion, not fruit. Audio fabricated eating sounds; vision misidentified onion as fruit.}}
\end{tcolorbox}
\end{minipage}%
\hfill
\begin{minipage}[t]{0.48\textwidth}
\begin{tcolorbox}[colback=purple!5,colframe=purple!50!black,title={\scriptsize Merger: ``What are you calling?'' \textbf{GT: A (gibbon)}},fonttitle=\scriptsize\bfseries,left=2pt,right=2pt,top=2pt,bottom=2pt]
\centering
\includegraphics[width=0.55\linewidth]{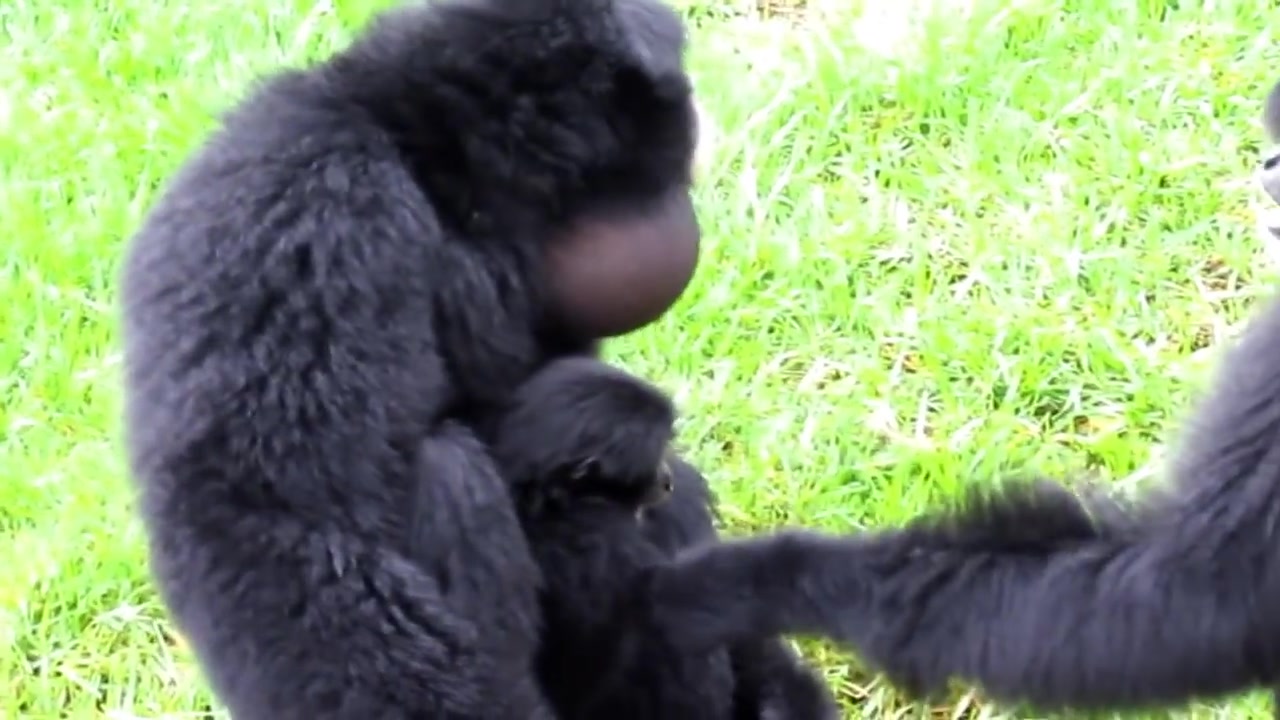}\\[0.3em]
\scriptsize\raggedright
\textbf{Pred: A} \checkmark\\[0.2em]
\textbf{Audio:} Correctly identifies gibbon vocalizations.\\[0.2em]
\textbf{Vision:} Identifies a primate with dark fur (but incorrectly claims a prehensile tail).\\[0.2em]
\textbf{Merged:} ``This creature is seen \underline{swinging from branches}\ldots the calls are unmistakably those of a gibbon.''\\[0.2em]
\textit{\textcolor{gray}{Judge: Gibbon is sitting on grass, never swinging. Merger fabricated this---neither teacher mentioned it.}}
\end{tcolorbox}
\end{minipage}

\caption{Qualitative examples of hallucination sources in the AVRT pipeline. Each box represents a different source category: \textcolor{blue!50!black}{\textbf{Audio Teacher}} (top-left), the audio teacher fabricated bee buzzing sounds not present in the audio; \textcolor{red!50!black}{\textbf{Vision Teacher}} (top-right), the vision teacher missed a visible sail and claimed the person was paddling; \textcolor{orange!50!black}{\textbf{Both Teachers}} (bottom-left), the audio teacher fabricated eating sounds and the vision teacher misidentified a raw onion as fruit; \textcolor{purple!50!black}{\textbf{Merger}} (bottom-right), the merger invented branch-swinging behavior absent from both teacher traces. Hallucinated claims are \underline{underlined}. Despite hallucinations, all examples reach the correct answer.}
\label{fig:halluc_source_examples}
\end{figure*}

\noindent\textbf{Hallucination Source Analysis.}
To understand the origins of hallucinations in the AVRT pipeline, we conduct a follow-up analysis on the 66 traces (11.1\%) flagged as hallucinating. Using Gemini-3.1-Flash-Lite as a judge, we present the original video and audio alongside the individual teacher traces and the final merged trace, and ask the model to attribute each hallucinated claim to its source: the audio teacher, the vision teacher, the merger model, or both teachers jointly (see Figure~\ref{fig:halluc_source_prompt} for the full prompt). The majority of hallucinations originate from both teachers contributing false claims (6.7\% of all samples), indicating that cross-modal consistency issues in the teacher outputs are the primary driver. Audio-only and vision-only teacher hallucinations account for 1.7\% and 1.5\% respectively. Notably, the merger introduces novel hallucinations in only 1.2\% of cases, confirming that it is largely faithful to its inputs and that improving individual teacher accuracy would address most hallucination issues. Figure~\ref{fig:halluc_source_examples} presents qualitative examples from each hallucination source category.

\begin{table}[t]
\centering
\small
\setlength{\tabcolsep}{4pt}
\begin{tabular}{l cc}
\toprule
\textbf{Metric} & \textbf{AVRT} & \textbf{Qwen3-Omni} \\
\midrule
Accurate          & \textbf{88.5\%} & 87.1\% \\
Hallucinating     & 11.1\% & 10.9\% \\
\rowcolor{gray!8}
\multicolumn{3}{l}{\quad\smash{\raisebox{0.3em}{\scriptsize$\hookrightarrow$}}\;\textit{\scriptsize Hallucination source (AVRT):}} \\
\rowcolor{gray!8}
\quad Both teachers      & \multicolumn{1}{r}{\scriptsize 6.7\%} & \\
\rowcolor{gray!8}
\quad Audio teacher only & \multicolumn{1}{r}{\scriptsize 1.7\%} & \\
\rowcolor{gray!8}
\quad Vision teacher only & \multicolumn{1}{r}{\scriptsize 1.5\%} & \\
\rowcolor{gray!8}
\quad Merger introduced  & \multicolumn{1}{r}{\scriptsize 1.2\%} & \\
Missing Info      & \textbf{0.3\%}  &  2.0\% \\
\cmidrule{1-3}
Audio Grounded    & \textbf{93.4\%} & 81.3\% \\
Visual Grounded   & 93.4\% & \textbf{94.6\%} \\
\bottomrule
\end{tabular}
\caption{\textbf{Individual trace evaluation on ${\sim}600$ AVRT-20K validation samples.}
Both models achieve comparable accuracy (${\sim}88\%$) and hallucination rates (${\sim}11\%$). AVRT achieves substantially higher audio grounding (93.4\% vs.\ 81.3\%) and lower missing-info rate (0.3\% vs.\ 2.0\%). Best results in \textbf{bold}. The gray sub-rows decompose AVRT's 11.1\% hallucination rate by source; the merger introduces novel hallucinations in only 1.2\% of cases.}
\label{tab:trace_quality}
\label{tab:halluc_source}
\end{table}

\begin{figure*}[t]
\centering
\small

\begin{minipage}[t]{0.48\textwidth}
\begin{tcolorbox}[colback=green!5,colframe=green!50!black,title={\scriptsize Accurate: ``What are the people doing?'' \textbf{GT: D (sew clothes)}},fonttitle=\scriptsize\bfseries,left=2pt,right=2pt,top=2pt,bottom=2pt]
\centering
\includegraphics[width=0.6\linewidth]{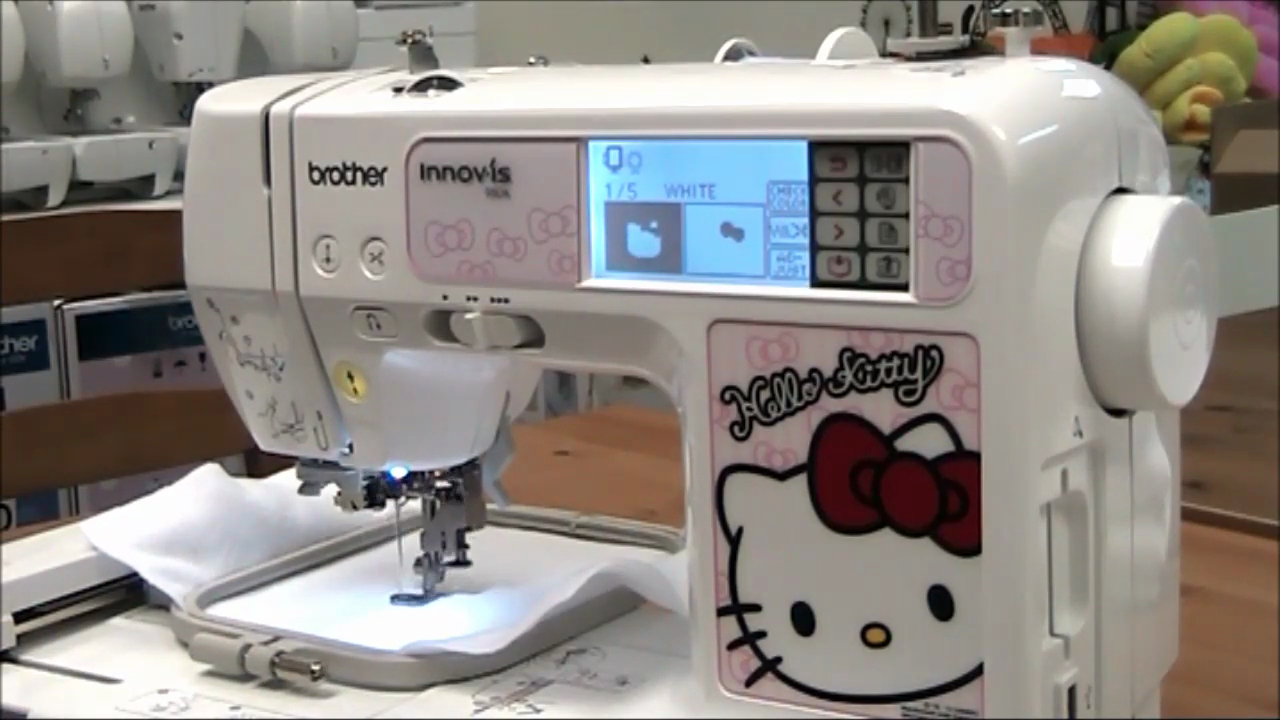}\\[0.3em]
\scriptsize
\textbf{Pred: D} \checkmark\\[0.2em]
The \textcolor{blue}{audio} clearly captures the rhythmic sounds of a sewing machine in operation. \textcolor{red}{Visually}, the video displays a Brother Innov-is sewing machine with Hello Kitty branding\ldots the needle is visible piercing through fabric held by the presser foot\ldots
\end{tcolorbox}
\end{minipage}%
\hfill
\begin{minipage}[t]{0.48\textwidth}
\begin{tcolorbox}[colback=green!5,colframe=green!50!black,title={\scriptsize Accurate: ``What happened in the video?'' \textbf{GT: D (hail)}},fonttitle=\scriptsize\bfseries,left=2pt,right=2pt,top=2pt,bottom=2pt]
\centering
\includegraphics[width=0.6\linewidth]{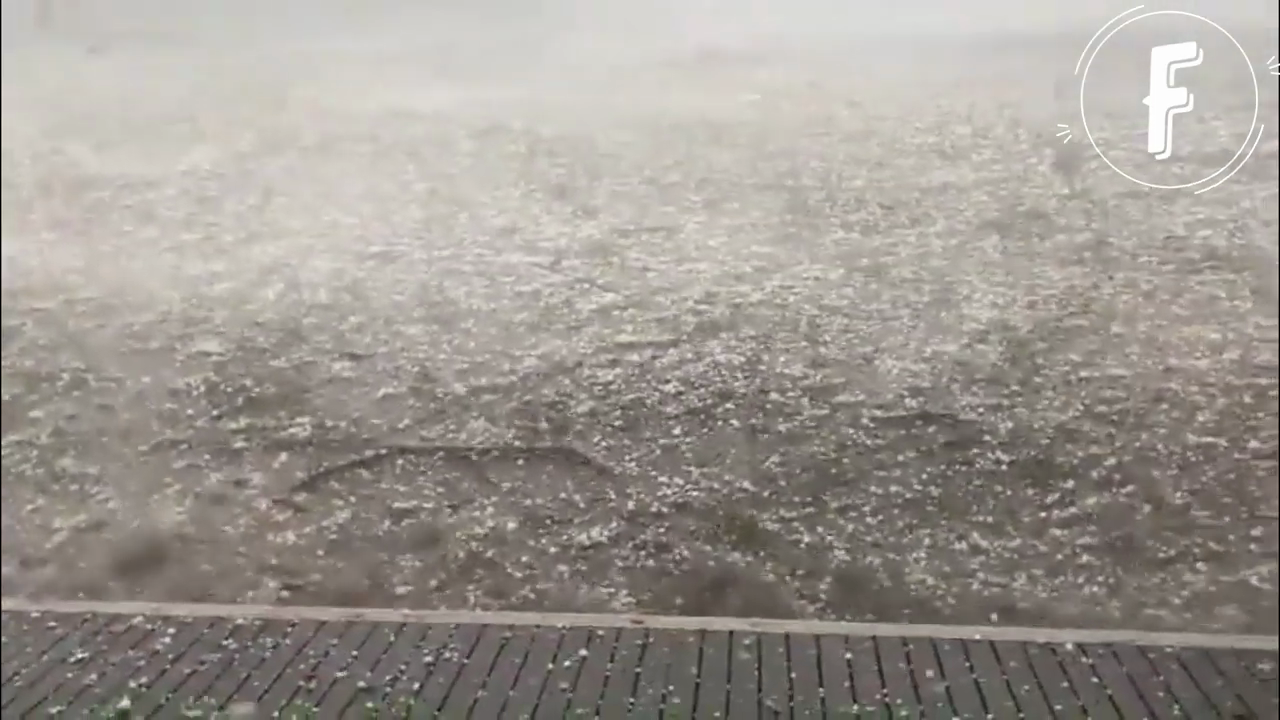}\\[0.3em]
\scriptsize
\textbf{Pred: D} \checkmark\\[0.2em]
The \textcolor{red}{visual} component shows small white round objects on the ground\ldots The \textcolor{blue}{audio} captures heavy rainfall along with distinct sounds of hail hitting surfaces\ldots intermittent shouts in the background might be people reacting to the hailstorm\ldots
\end{tcolorbox}
\end{minipage}

\begin{minipage}[t]{0.48\textwidth}
\begin{tcolorbox}[colback=green!5,colframe=green!50!black,title={\scriptsize Accurate: ``What's driving in the video?'' \textbf{GT: D (train)}},fonttitle=\scriptsize\bfseries,left=2pt,right=2pt,top=2pt,bottom=2pt]
\centering
\includegraphics[width=0.6\linewidth]{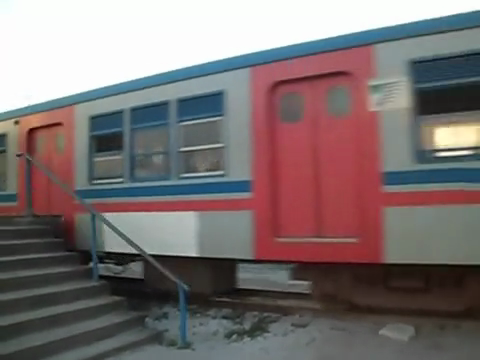}\\[0.3em]
\scriptsize
\textbf{Pred: D} \checkmark\\[0.2em]
The \textcolor{red}{video} displays a large, multi-sectioned vehicle that moves along railroad tracks with people boarding through a staircase. The \textcolor{blue}{audio} clearly emits the sounds associated with a train, such as the rhythmic clacking of wheels and the deep rumble of its engine\ldots
\end{tcolorbox}
\end{minipage}%
\hfill
\begin{minipage}[t]{0.48\textwidth}
\begin{tcolorbox}[colback=yellow!5,colframe=yellow!50!black,title={\scriptsize Missing Info: ``What is the main source of sound?'' \textbf{GT: D (civil defense alarm)}},fonttitle=\scriptsize\bfseries,left=2pt,right=2pt,top=2pt,bottom=2pt]
\centering
\includegraphics[width=0.6\linewidth]{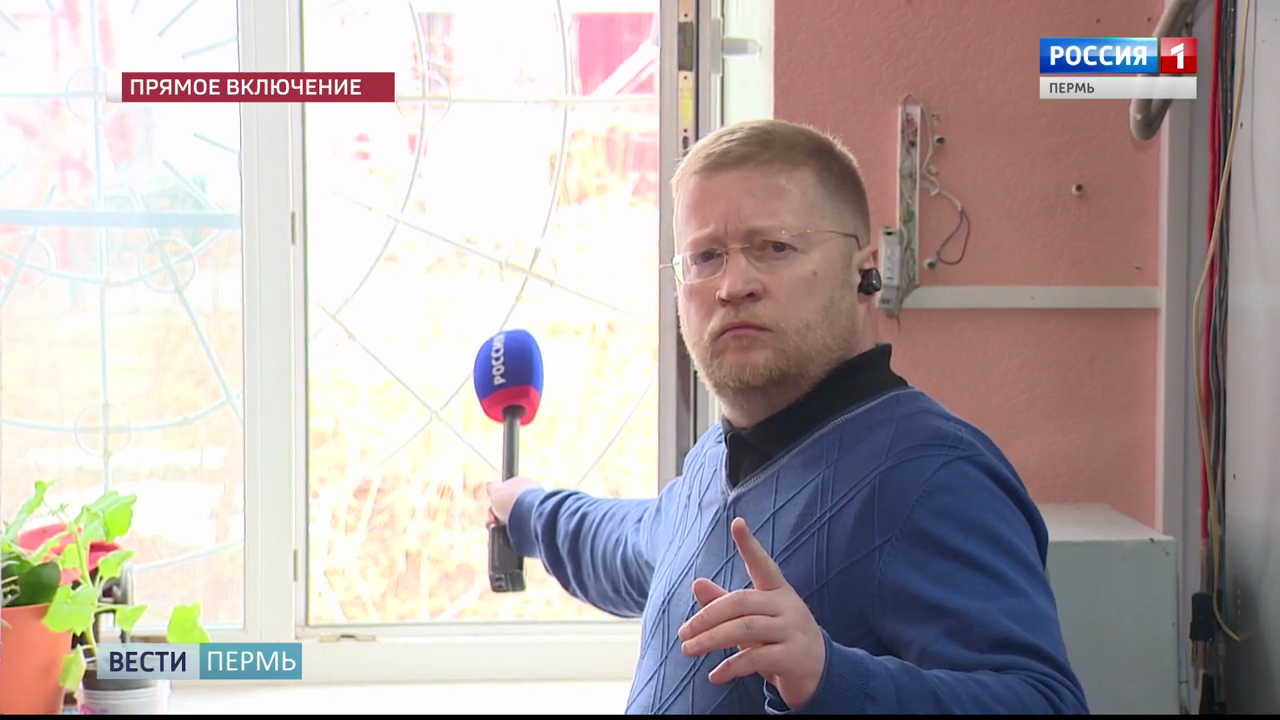}\\[0.3em]
\scriptsize
\textbf{Pred: D} \checkmark\\[0.2em]
The \textcolor{red}{image} depicts a news report with the logo of ``Rossiya 1'' indicating a live broadcast\ldots the most plausible sound for a news broadcast discussing civil defense is the civil defense alarm\ldots
\\[0.2em]
\textit{\textcolor{gray}{Judge: Correct answer, but fails to describe the actual alarm sound heard in the audio.}}
\end{tcolorbox}
\end{minipage}

\begin{minipage}[t]{0.48\textwidth}
\begin{tcolorbox}[colback=red!5,colframe=red!50!black,title={\scriptsize Hallucinating: ``What are the people doing?'' \textbf{GT: B (rowboat)}},fonttitle=\scriptsize\bfseries,left=2pt,right=2pt,top=2pt,bottom=2pt]
\centering
\includegraphics[width=0.6\linewidth]{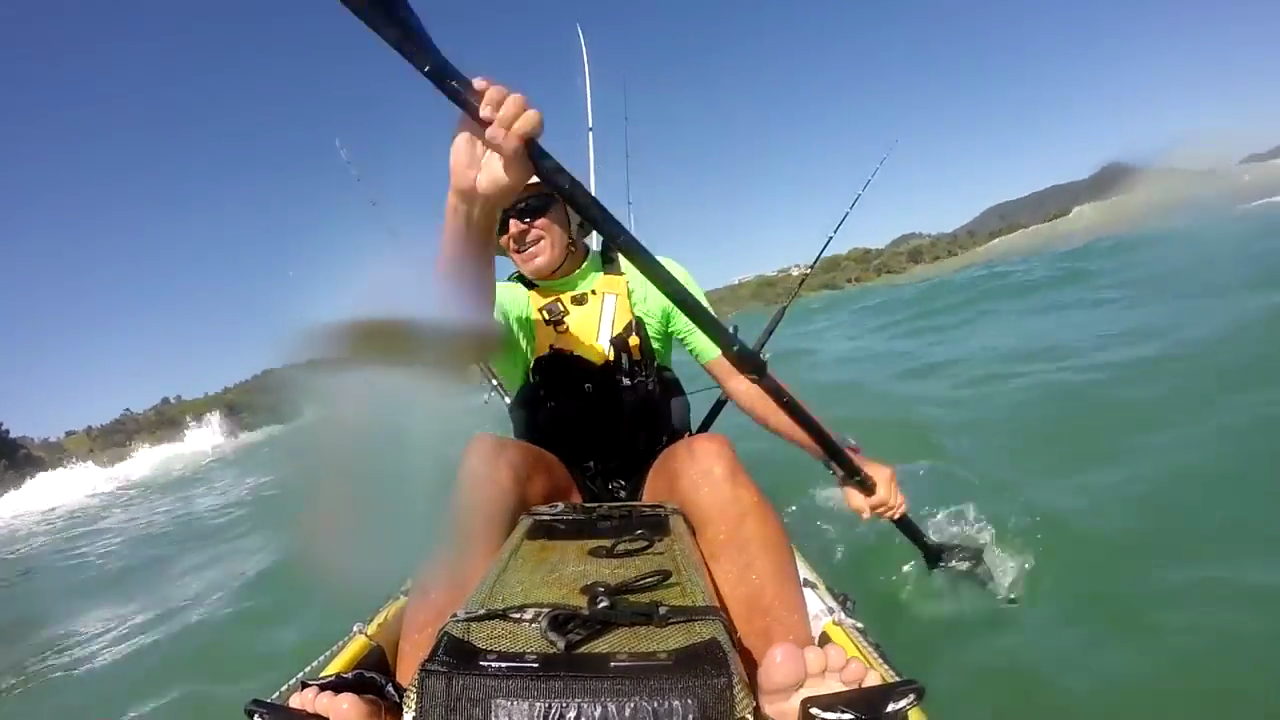}\\[0.3em]
\scriptsize
\textbf{Pred: B} \checkmark\\[0.2em]
The \textcolor{blue}{audio} has clear sounds of water being displaced by \underline{oars}\ldots \textcolor{red}{visuals} showing a person maneuvering a small boat using \underline{two paddles}\ldots
\\[0.2em]
\textit{\textcolor{gray}{Judge: Video shows a kayak with a single double-bladed paddle, not oars or two paddles.}}
\end{tcolorbox}
\end{minipage}%
\hfill
\begin{minipage}[t]{0.48\textwidth}
\begin{tcolorbox}[colback=red!5,colframe=red!50!black,title={\scriptsize Hallucinating: ``Where does this video take place?'' \textbf{GT: D (in the woods)}},fonttitle=\scriptsize\bfseries,left=2pt,right=2pt,top=2pt,bottom=2pt]
\centering
\includegraphics[width=0.6\linewidth]{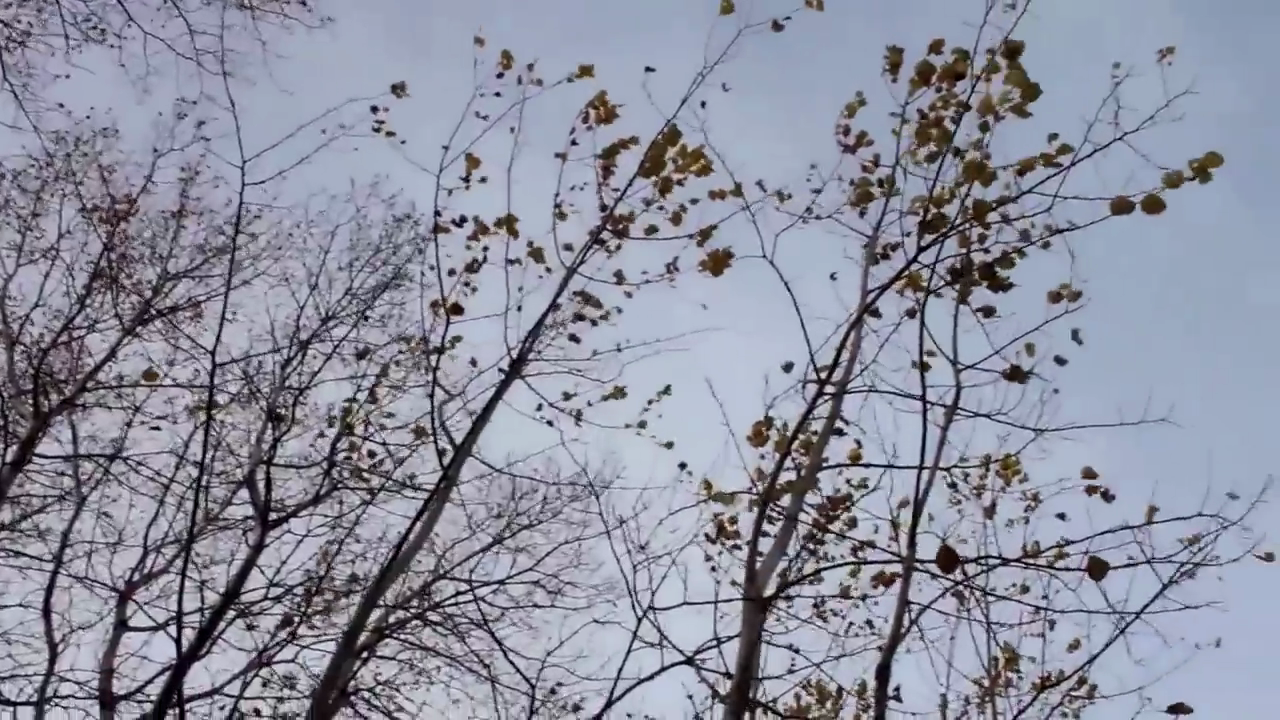}\\[0.3em]
\scriptsize
\textbf{Pred: D} \checkmark\\[0.2em]
The \textcolor{red}{video} showcases tall, thin deciduous trees that appear bare\ldots the \textcolor{blue}{audio} provides a significant clue: there is a noticeable sound of a \underline{distant waterfall}\ldots
\\[0.2em]
\textit{\textcolor{gray}{Judge: Correctly identifies forest visually, but hallucinates the waterfall sound.}}
\end{tcolorbox}
\end{minipage}

\caption{Qualitative examples from the automated reasoning trace quality evaluation. \textcolor{green!50!black}{\textbf{Green}}: accurate traces with correct cross-modal reasoning. \textcolor{red!50!black}{\textbf{Red}}: traces with hallucinated details (\underline{underlined}), though still reaching the correct answer. \textcolor{yellow!50!black}{\textbf{Yellow}}: correct answer but missing explicit audio analysis. Color highlights in traces: \textcolor{blue}{audio} and \textcolor{red}{visual} evidence.}
\label{fig:quality_examples}
\end{figure*}

Figure~\ref{fig:quality_examples} presents qualitative examples from each category. The \textit{Accurate} examples demonstrate genuine cross-modal reasoning: the sewing example integrates the rhythmic machine sounds with the visual identification of a Hello Kitty sewing machine, while the hail example combines visual observation of white objects on the ground with the characteristic audio of hail striking surfaces. The \textit{Hallucinating} examples illustrate typical reasoning errors: the kayak trace inaccurately describes ``two paddles'' and ``oars'' instead of a single double-bladed paddle, while the forest trace fabricates a ``distant waterfall'' sound despite correctly identifying the woodland setting visually. The \textit{Missing Info} example shows the model correctly identifying a civil defense alarm from visual context (a Russian news broadcast) but failing to describe the actual alarm sound in its reasoning.

\begin{figure*}[t]
\begin{tcolorbox}[colback=gray!10,colframe=gray!50,title=Head-to-Head Trace Comparison Prompt (Gemini-3.1-Flash-Lite)]
\small\ttfamily
You are an expert judge comparing two audio-visual reasoning traces.

You will be given:
1. A video file (with its original audio track)
2. A separate audio file extracted from the same video
3. A multiple-choice question about the video
4. The answer choices
5. The ground-truth correct answer
6. Two reasoning traces (Trace A and Trace B) generated by different AI systems

Your task is to carefully watch the video, listen to the audio, read both reasoning traces, and determine which trace is better based on these criteria:

\textbf{Factual Accuracy} --- Are the observations about the video and audio content correct?

\textbf{Completeness} --- Does the trace address both audio and visual evidence?

\textbf{Logical Soundness} --- Is the reasoning chain coherent and well-structured?

\textbf{Groundedness} --- Are claims about audio and visual content supported by the actual media?

[Question, Answer Choices, Ground-Truth Answer, Trace A, and Trace B are provided]

Respond with a JSON object with fields: ``winner'' (``A'', ``B'', or ``Tie''), ``confidence'' (0.0--1.0), ``explanation'', ``trace\_a\_verdict'' (ACCURATE / MISSING\_INFO / HALLUCINATING), ``trace\_b\_verdict'', ``trace\_a\_audio\_grounded'' (boolean), ``trace\_a\_visual\_grounded'' (boolean), ``trace\_b\_audio\_grounded'' (boolean), ``trace\_b\_visual\_grounded'' (boolean).
\end{tcolorbox}
\caption{The prompt used for head-to-head comparison of reasoning traces. The judge model (Gemini-3.1-Flash-Lite) receives the original video, audio, and two reasoning traces from different systems (AVRT and Qwen3-Omni-Thinking, with randomized assignment to A/B positions). It selects a winner based on factual accuracy, completeness, logical soundness, and groundedness, while also returning per-trace quality verdicts and per-modality grounding flags.}
\label{fig:quality_eval_prompt}
\end{figure*}

\begin{figure*}[t]
\begin{tcolorbox}[colback=gray!10,colframe=gray!50,title=Hallucination Source Analysis Prompt (Gemini-3.1-Flash-Lite)]
\small\ttfamily
You are an expert judge performing a detailed hallucination source analysis on an audio-visual reasoning trace.

An AI system produced a final reasoning trace by merging two teacher traces: (1) an \textbf{audio teacher} that analyzed only the audio, and (2) a \textbf{vision teacher} that analyzed only the video frames. A \textbf{merger model} then combined these into a single unified reasoning trace. A previous evaluation flagged the final merged trace as HALLUCINATING. Your job is to determine WHERE the hallucination originates.

You are given: a video file (with audio), a separate audio file, a multiple-choice question with answer choices, the ground-truth answer, the audio teacher's reasoning, the vision teacher's reasoning, the final merged trace, and the previous judge's explanation.
\\[1em]
\textbf{Categories for hallucination source:}\\[1em]
\textbf{AUDIO\_TEACHER} --- The hallucinated content traces back to a false claim in the audio teacher's reasoning.\\[1em]
\textbf{VISION\_TEACHER} --- The hallucinated content traces back to a false claim in the vision teacher's reasoning.\\[1em]
\textbf{MERGER} --- The hallucinated content was NOT present in either teacher trace. The merger model invented or distorted information.\\[1em]
\textbf{BOTH\_TEACHERS} --- Both teachers contributed hallucinated content that ended up in the final trace.\\[1em]

[Question, Answer Choices, Ground-Truth Answer, Audio Teacher's Reasoning, Vision Teacher's Reasoning, Final Merged Trace, and Previous Evaluation Context are provided]

Respond with a JSON object with fields: ``hallucination\_source'' (AUDIO\_TEACHER, VISION\_TEACHER, MERGER, or BOTH\_TEACHERS), ``confidence'' (0.0--1.0), ``hallucinated\_claims'' (list of false claims), ``source\_per\_claim'' (source for each claim), ``audio\_teacher\_accurate'' (boolean), ``vision\_teacher\_accurate'' (boolean), ``merger\_faithful'' (boolean), ``explanation'' (2--4 sentences).
\end{tcolorbox}
\caption{The prompt used for hallucination source analysis. The judge model (Gemini-3.1-Flash-Lite) receives the original video, audio, both teacher traces, and the merged trace flagged as hallucinating. It attributes each hallucinated claim to its source: the audio teacher, vision teacher, merger, or both teachers.}
\label{fig:halluc_source_prompt}
\end{figure*}

\subsection{Model Behavior Under Missing Modalities} \label{app:modality_consistency}

To investigate whether AVRT hallucinates modality-specific information when one modality is unavailable, we evaluate the model with masked inputs. We test the model in three conditions: (1) both audio and visual inputs provided (standard setting), (2) only video frames with silent audio, and (3) only audio with blank video frames. This experiment examines whether the model acknowledges missing information or fabricates details about the absent modality. The evaluation uses the same OmniBench test set.

\begin{figure*}[t]
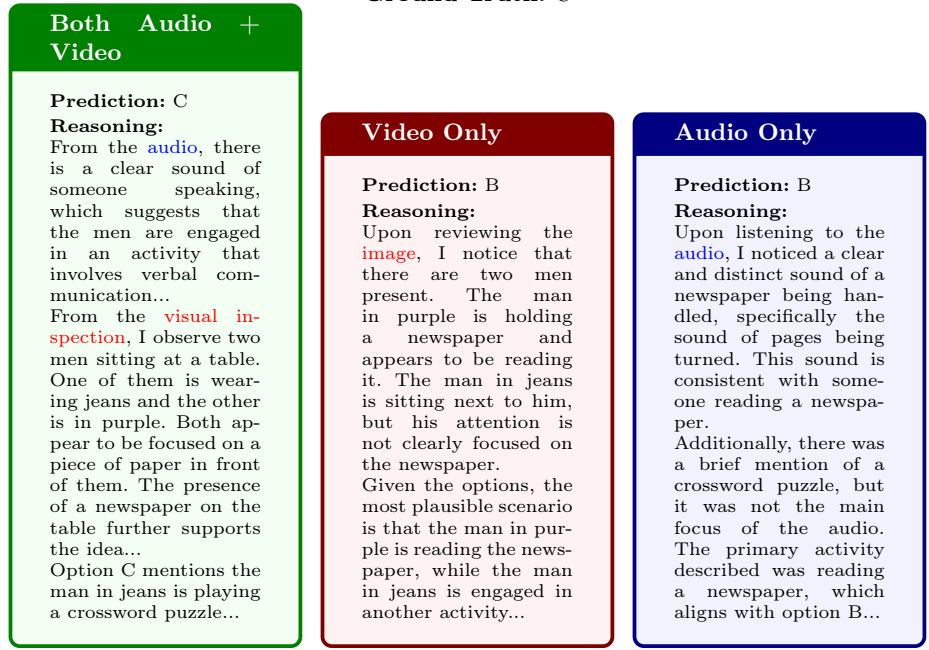

\centering
\small
\textbf{Question:} What are the men doing?\\
\textbf{Options:} (A) The man in jeans is taking notes from the newspaper. (B) The man in purple is reading the newspaper. (C) The man in jeans is playing a crossword puzzle. (D) The man on the table is doing a crossword puzzle.\\
\textbf{Ground Truth:} C

\begin{minipage}[t]{0.32\textwidth}
\begin{tcolorbox}[colback=green!5,colframe=green!50!black,title=Both Audio + Video,fonttitle=\small\bfseries]
\scriptsize
\textbf{Prediction:} C \\[0.2em]
\textbf{Reasoning:}\\
From the \textcolor{blue}{audio}, there is a clear sound of someone speaking, which suggests that the men are engaged in an activity that involves verbal communication...

From the \textcolor{red}{visual inspection}, I observe two men sitting at a table. One of them is wearing jeans and the other is in purple. Both appear to be focused on a piece of paper in front of them. The presence of a newspaper on the table further supports the idea...

Option C mentions the man in jeans is playing a crossword puzzle...
\end{tcolorbox}
\end{minipage}%
\hfill
\begin{minipage}[t]{0.32\textwidth}
\begin{tcolorbox}[colback=red!5,colframe=red!50!black,title=Video Only,fonttitle=\small\bfseries]
\scriptsize
\textbf{Prediction:} B \\[0.2em]
\textbf{Reasoning:}\\
Upon reviewing the \textcolor{red}{image}, I notice that there are two men present. The man in purple is holding a newspaper and appears to be reading it. The man in jeans is sitting next to him, but his attention is not clearly focused on the newspaper.

Given the options, the most plausible scenario is that the man in purple is reading the newspaper, while the man in jeans is engaged in another activity...
\end{tcolorbox}
\end{minipage}%
\hfill
\begin{minipage}[t]{0.32\textwidth}
\begin{tcolorbox}[colback=blue!5,colframe=blue!50!black,title=Audio Only,fonttitle=\small\bfseries]
\scriptsize
\textbf{Prediction:} B \\[0.2em]
\textbf{Reasoning:}\\
Upon listening to the \textcolor{blue}{audio}, I noticed a clear and distinct sound of a newspaper being handled, specifically the sound of pages being turned. This sound is consistent with someone reading a newspaper.

Additionally, there was a brief mention of a crossword puzzle, but it was not the main focus of the audio. The primary activity described was reading a newspaper, which aligns with option B...
\end{tcolorbox}
\end{minipage}

\caption{Example of model reasoning traces when different modalities are masked. With both modalities (left), the model integrates audio and visual evidence. With video only (center), reasoning focuses on visual observations without hallucinating audio content. With audio only (right), the model constrains itself to acoustic features without fabricating visual details. Color highlights indicate modality-specific reasoning: \textcolor{blue}{audio} and \textcolor{red}{visual}.}
\label{fig:modality_masking}
\end{figure*}

Table~\ref{tab:ablation_single_modality} in the main paper shows that performance degrades by 14.4 points when only audio is available and by 10.5 points when only video is available, demonstrating that AVRT appropriately relies on both modalities. Figure~\ref{fig:modality_masking} provides a qualitative example comparing reasoning traces across three conditions. When both modalities are provided, the model explicitly integrates evidence using phrases like ``From the audio...'' and ``From the visual inspection...'' to arrive at the correct answer. When only video is available, the model constrains reasoning to visual observations (``Upon reviewing the image...'') without fabricating audio content, though this leads to an incorrect prediction due to missing acoustic cues. Similarly, with audio-only input, the model focuses exclusively on sounds without hallucinating visual details like clothing or positions. While encouraging, this behavior could be sensitive to prompting strategies. The model may occasionally reference missing modalities due to strong pretraining priors.

\subsection{Control Experiments: Role of Reasoning Content} \label{app:control_experiments}

To validate that AVRT's improvements stem from the actual reasoning content in the composed traces, rather than from additional training data, learning the output format, or simply seeing correct answers, we conduct a series of control experiments. Each control ablates the SFT data composition while keeping all other training hyperparameters fixed, isolating the contribution of specific pipeline components.

We evaluate three controls against the baseline and the full AVRT pipeline:
\begin{enumerate}
    \item \textit{Answer-only SFT}: the model is fine-tuned on the AVRT-20K training samples containing only the correct answer letter (\eg ``A''), without reasoning traces and without subsequent RL.
    \item \textit{Answer-only SFT + RL}: same answer-only SFT followed by RL, testing whether RL can compensate for the lack of reasoning supervision.
    \item \textit{Format-only SFT + RL}: SFT on traces with the correct \texttt{<think>}\ldots\texttt{</think>}\allowbreak\texttt{<answer>}\ldots\texttt{</answer>} structure and the correct answer letter, but generic, non-informative filler text (\eg ``Let me analyze this question carefully\ldots'') instead of actual cross-modal reasoning, followed by RL.
\end{enumerate}

\begin{table}[t]
\centering
\small
\setlength{\tabcolsep}{4pt}
\resizebox{\linewidth}{!}{%
\begin{tabular}{lcccc}
\toprule
\textbf{Configuration} & \textbf{RL} & \textbf{OmniBench} & \textbf{DailyOmni} & \textbf{MMAR} \\
\midrule
Baseline (Qwen2.5-Omni-3B) & $\times$ & 50.2 & 43.1 & 53.7 \\
\midrule
Answer-only SFT & $\times$ & 46.0 \scriptsize{\textcolor{red}{($-$4.2)}} & 47.1 \scriptsize{\textcolor{darkgreen}{(+4.0)}} & 38.0 \scriptsize{\textcolor{red}{($-$15.7)}} \\
Answer-only SFT + RL & \checkmark & 12.5 \scriptsize{\textcolor{red}{($-$37.7)}} & 28.9 \scriptsize{\textcolor{red}{($-$14.2)}} & 45.2 \scriptsize{\textcolor{red}{($-$8.5)}} \\
Format-only SFT + RL$^\dagger$ & \checkmark & 21.3 \scriptsize{\textcolor{red}{($-$28.9)}} & 29.9 \scriptsize{\textcolor{red}{($-$13.2)}} & 22.1 \scriptsize{\textcolor{red}{($-$31.6)}} \\
\midrule
\rowcolor{almond}\textbf{AVRT (Ours)} & \checkmark & \textbf{56.3} \scriptsize{\textcolor{darkgreen}{(+6.1)}} & \textbf{49.2} \scriptsize{\textcolor{darkgreen}{(+6.1)}} & \textbf{57.3} \scriptsize{\textcolor{darkgreen}{(+3.6)}} \\
\bottomrule
\end{tabular}%
}
\caption{\textbf{Control experiments isolating the role of reasoning trace content.} All models use Qwen2.5-Omni-3B as the base. ``Answer-only SFT'' trains on 20K samples with correct answer letters only (no reasoning traces). $^\dagger$``Format-only'' uses \texttt{<think>}generic filler\texttt{</think><answer>}correct letter\texttt{</answer>} traces that match AVRT's output structure but contain no informative reasoning. Deltas are relative to the baseline. Neither answer-only nor format-only training approaches AVRT, confirming that the actual reasoning content in the composed traces is the key driver of performance.}
\label{tab:control_experiments}
\end{table}

Table~\ref{tab:control_experiments} summarizes the results across three benchmarks.

\noindent\textbf{Answer-only SFT.} The answer-only SFT model produces clean single-letter outputs (99.6\% well-formed on OmniBench) but accuracy degrades versus the baseline, particularly on MMAR ($-15.7$ points). The model learns to map inputs directly to answer letters in the AVQA domain, bypassing any reasoning process, which leads to superficial pattern matching that does not consistently generalize to other benchmarks.

\noindent\textbf{Answer-only SFT + RL: output collapse.} Adding RL on top of the answer-only SFT leads to severe output collapse. On OmniBench, accuracy drops to 12.5\%, with 48\% of outputs being invalid. This collapse arises from a conflict between the SFT initialization and the RL objective: the SFT stage trains the model to output bare answer letters (\eg ``B''), but the RL format reward pushes toward structured \texttt{<think>}\allowbreak\texttt{...}\allowbreak\texttt{</think>}\allowbreak\texttt{<answer>}\allowbreak\texttt{...}\allowbreak\texttt{</answer>} outputs. The model attempts to adopt the tag structure but, lacking any prior exposure to reasoning traces, cannot produce coherent content within it. Among the 52\% of valid outputs, most are empty thinking blocks (\eg \texttt{<think>}\allowbreak\texttt{</think>}\allowbreak\texttt{<answer>}\allowbreak\texttt{D}\allowbreak\texttt{</answer>}), where the model learned the tags but has nothing to put inside them. The remaining 48\% are broken fragments: 67\% of invalid outputs consist of an orphaned \texttt{</answer>} closing tag with no content, 29\% start with \texttt{<think>} but never close properly, and 4\% are sequences of closing tags only (\eg \texttt{</think>}\allowbreak\texttt{</answer>}). Even among valid outputs, accuracy is poor because the model never learned cross-modal reasoning patterns. The model receives conflicting training signals: SFT encourages bare answer letters while RL rewards structured tag formatting.

\noindent\textbf{Format-only SFT + RL: template reasoning.} The format-only control exhibits a different but similarly detrimental failure mode. Unlike the answer-only variant, this model produces well-structured, plausible-looking outputs: 85\% of OmniBench responses are validly formatted with numbered reasoning steps and modality-by-modality analysis. However, the content is generic template filler rather than genuine observation. A typical output follows the pattern: \textit{``The audio does not provide any specific information about the men's actions. The video shows two men sitting at a table. Based on the available evidence\ldots''} The model learned the \textit{form} of reasoning perfectly (structured steps, modality labels, concluding statements) but applies the same template regardless of the actual audio-visual content. The 15\% of invalid outputs follow the same template structure but are truncated before the closing \texttt{</answer>} tag, suggesting the model generates increasingly verbose filler that exceeds the maximum output length.

This template reasoning explains the progressive performance collapse during RL training. Using the same experimental setup as the main paper, the accuracy reward signal is too noisy to learn from (since the model is not actually reasoning about the content), while the format and length rewards provide consistent positive signal for producing longer, well-structured text. The model thus over-optimizes for surface-level reward components while accuracy degrades to 21.3\% on OmniBench and 22.1\% on MMAR.

\noindent\textbf{Implications.} AVRT achieves 56.3\% on OmniBench, 49.2\% on DailyOmni, and 57.3\% on MMAR, outperforming all controls by a large margin. The contrast between AVRT and the format-only control is particularly informative: both use the same output format, but only AVRT contains genuine cross-modal reasoning. None of the alternative SFT configurations could replicate the performance gains observed with AVRT, suggesting that the actual reasoning content in the composed traces is the key driver. In our setting, RL alone was not sufficient to compensate for the absence of reasoning supervision during SFT: without that foundation, RL either collapsed the model's outputs (answer-only) or optimized surface features while degrading accuracy (format-only).

\section{AVRT Additional Information} \label{app:additional_info}

\subsection{AVRT Dataset Construction Prompts} \label{app:prompts}
In this section, we provide the prompts used for both teacher models (Kimi-VL-Thinking and Audio Flamingo 3 (\textit{think})), and the merger model (Qwen2.5-14B-Instruct).

\noindent\textbf{Visual Teacher Prompt.}
We design the visual teacher prompt to elicit detailed temporal reasoning from video frames. As shown in Figure~\ref{fig:visual_teacher_prompt}, the prompt instructs Kimi-VL-Thinking to analyze 8 evenly distributed frames from each video, providing both a comprehensive visual description and explicit reasoning that considers temporal progression. The prompt includes a concrete example that demonstrates the expected response format, encouraging the model to describe what it observes across frames, reason about the visual evidence, and arrive at the correct answer. For questions requiring straightforward visual identification, the prompt allows for brief reasoning while still maintaining the descriptive component.

\begin{figure*}[t]
\begin{tcolorbox}[colback=gray!10,colframe=gray!50,title=Visual Teacher (Kimi-VL-Thinking) Prompt]
\small\ttfamily
You are an intelligent vision agent. I will provide you with 8 representative frames from a video (evenly distributed across the video duration) and a question about the video content in MCQ format. You need to first provide a thorough description of what you're seeing across these video frames, then add Chain-of-Thought-type reasoning to analyze the visual content, and finally provide your answer. Here is an example:

Input Question: What type of activity is happening in this video? Choose one among the following options:(A) Crime thriller scene (B) Documentary narration\textbackslash n(C) Romantic comedy scene\textbackslash n(D) Action movie or racing scene\textbackslash n

Expected response format:\textbackslash n

Visual Description: Across these video frames, I can see a progression of high-speed chase scenes with vehicles moving rapidly through an urban environment. The frames show consistent dynamic motion, intense lighting, and what appears to be an ongoing action sequence with cars and possibly motorcycles. The temporal progression across frames reveals the continuous high-energy nature of the content.\textbackslash n

Reasoning: Based on the consistent high-speed vehicle movement visible across multiple frames, the sustained dynamic camera work, intense lighting throughout the sequence, and the overall action-oriented visual elements that persist across the video timeline, this content would be most suitable for action-focused scenarios that require high-energy sequences. The visual elements strongly suggest this is an action movie or racing scene rather than other genres like crime thriller, documentary, or romantic comedy.\textbackslash n

Answer: (D) Action movie or racing scene\textbackslash n

Follow this format: provide a detailed visual description analyzing the temporal progression across frames, then your reasoning considering the full video context and evaluating each option, then the final answer. For answers that do not require complex reasoning (e.g., for a question like "What color is the object?" or "How many people are in the image?" where the answer is direct), still provide the visual description but keep the reasoning brief.\textbackslash n

Here is the input question:
\end{tcolorbox}
\caption{The prompt used for the visual teacher model (Kimi-VL-Thinking). The prompt guides the model to provide temporal visual analysis across video frames, followed by explicit reasoning and a final answer.}
\label{fig:visual_teacher_prompt}
\end{figure*}

\noindent\textbf{Audio Teacher Prompt.}
For the audio teacher, we adopt the prompts from the Audio Flamingo 3 paper to ensure optimal performance and maintain consistency with the original model's training methodology. As illustrated in Figure~\ref{fig:audio_teacher_prompt}, the audio prompt follows a similar structure to the visual prompt, instructing the model to provide a thorough audio description, evaluate each answer option with explicit reasoning, and then provide the final answer. The prompt includes an example that demonstrates how to analyze audio characteristics (e.g., beat, vocals, production style) and map them to appropriate contexts. This approach ensures that the audio teacher generates reasoning traces that complement the visual analysis.

\begin{figure*}[t]
\begin{tcolorbox}[colback=gray!10,colframe=gray!50,title=Audio Teacher (Audio Flamingo 3 \textit{think}) Prompt]
\small\ttfamily
You are an intelligent audio agent. I will provide you with an audio and a question about the audio in MCQ format. You need to first provide a thorough description of what you're hearing in the audio, then add Chain-of-Thought-type reasoning to analyze the audio content and evaluate each option, and finally provide your answer. Here is an example:

Input Question: What type of soundtrack would this piece be most suitable for? Choose one among the following options:(A) Crime thriller movie (B) Documentary narration\textbackslash n(C) Romantic comedy movie\textbackslash n(D) Futuristic movie or car racing video game\textbackslash n

Expected response format:\textbackslash n

Audio Description: This audio features a high-energy electronic track with a driving beat, synthesized sounds, and confident rap vocals. The lyrics mention themes of speed and success, including phrases like 'living automatic' and references to new cars. The production has a modern, polished sound with heavy use of electronic elements.\textbackslash n

Reasoning: Based on the driving beat, confident rap vocals, mentions of speed and success, and overall high-energy modern production with electronic elements, this piece would be most suitable for high-octane, modern scenarios that require energetic background music. Evaluating the options: (A) Crime thriller movies typically use more suspenseful, darker soundtracks; (B) Documentary narration usually requires more neutral, informative background music; (C) Romantic comedy movies generally feature lighter, more melodic soundtracks; (D) Futuristic movies or car racing video games would benefit from exactly this type of high-energy electronic music with themes of speed and technology.\textbackslash n

Answer: (D) Futuristic movie or car racing video game\textbackslash n

Follow this format: provide a detailed audio description first, then your reasoning that evaluates each option, then the final answer. For answers that do not require complex reasoning (e.g., for a question like "Who performs the vocals in this song?" or "What primary instrument is featured in this piece?" where the answer is direct), still provide the audio description but keep the reasoning brief.\textbackslash n

Here is the input question:
\end{tcolorbox}
\caption{The prompt used for the audio teacher model (Audio Flamingo 3 \textit{think}). The prompt instructs the model to describe audio characteristics and provide explicit reasoning that evaluates each answer option.}
\label{fig:audio_teacher_prompt}
\end{figure*}

\noindent\textbf{Merger Prompt.}
The merger model guides the combination of reasoning traces into a unified multimodal analysis. Figure~\ref{fig:merger_prompt} shows the prompt used for Qwen2.5-14B-Instruct, which receives the question along with both the audio and visual analyses generated by the teacher models. The merger does not receive the ground-truth answer; it must derive the correct answer solely from the teacher traces. The merged reasoning is formatted within \texttt{\textless think\textgreater} tags, followed by the final answer in \texttt{\textless answer\textgreater} tags.

\begin{figure*}[t]
\begin{tcolorbox}[colback=gray!10,colframe=gray!50,title=Merger (Qwen2.5-14B-Instruct) Prompt]
\small\ttfamily
You are an intelligent multimodal agent. I will provide you with a question in MCQ format, along with separate audio and visual analyses from specialized models. Your task is to merge these analyses into a coherent reasoning chain that integrates both modalities to arrive at the correct answer.

Question: \{question\}\{formatted\_choices\}

Audio Analysis: \{audio\_reasoning\}

Visual Analysis: \{vision\_reasoning\}

Instructions: \\
- Format your merged reasoning inside \textless think\textgreater ... \textless /think\textgreater \\
- At the end, output your final answer (just the letter, e.g., A, B, C, or D) inside \textless answer\textgreater ... \textless /answer\textgreater \\
- Write sentences that integrate both audio and visual evidence \\
- Explain how the audio and visual clues work together to lead you to the conclusion \\
- Make the explanation thorough but succinct \\

Combined Analysis:
\end{tcolorbox}
\caption{The prompt used for the merger model (Qwen2.5-14B-Instruct). The prompt guides the model to integrate audio and visual analyses into a coherent multimodal reasoning trace formatted with \texttt{\textless think\textgreater} and \texttt{\textless answer\textgreater} tags.}
\label{fig:merger_prompt}
\end{figure*}

\subsection{Additional Dataset Statistics}
    \definecolor{trainblue}{HTML}{4C72B0}
    \definecolor{valorange}{HTML}{DD8452}
    \definecolor{pieA}{HTML}{4C72B0}
    \definecolor{pieB}{HTML}{DD8452}
    \definecolor{pieC}{HTML}{55A868}
    \definecolor{pieD}{HTML}{C44E52}
    \definecolor{modAV}{HTML}{4C72B0}
    \definecolor{modSound}{HTML}{DD8452}
    \definecolor{modVisual}{HTML}{55A868}

    \begin{figure*}[h]
    \centering
    \begin{tikzpicture}
    \begin{axis}[
        ybar,
        width=0.92\textwidth,
        height=4.5cm,
        bar width=10pt,
        title={\small\textbf{Question Type Distribution}},
        symbolic x coords={Which, {Come From}, Happening, Where, Why, Others},
        xtick=data,
        xticklabel style={font=\scriptsize},
        yticklabel style={font=\scriptsize},
        ylabel style={font=\scriptsize},
        ylabel={Percentage (\%)},
        ymin=0, ymax=55,
        legend style={
            at={(0.98,0.97)}, anchor=north east,
            font=\scriptsize, draw=none, fill=white, fill opacity=0.8, text opacity=1,
            legend columns=1, column sep=3pt,
        },
        enlarge x limits=0.1,
        nodes near coords,
        nodes near coords style={font=\tiny},
        every axis plot/.append style={fill opacity=0.9},
        axis line style={gray!70},
        major grid style={gray!20, dashed},
        ymajorgrids=true,
        xtick pos=bottom,
    ]
    \addplot[fill=trainblue, draw=trainblue!80] coordinates {(Which,45.2) ({Come From},30.9) (Happening,15.5) (Where,8.0) (Why,0.2) (Others,0.2)};
    \addplot[fill=valorange, draw=valorange!80] coordinates {(Which,45.7) ({Come From},29.8) (Happening,14.1) (Where,9.7) (Why,0.4) (Others,0.3)};
    \legend{Train, Val}
    \end{axis}
    \end{tikzpicture}

    \begin{minipage}[t]{0.33\textwidth}
    \centering
    \resizebox{\textwidth}{!}{%
    \begin{tabular}{@{}c@{}}
    {\small\textbf{Answer Distribution}}\\[0.6em]
    \begin{tikzpicture}
    \pie[
        radius=1.6,
        text=pin,
        pin distance=0.4cm,
        color={pieA, pieB, pieC, pieD},
        before number=\scriptsize,
        after number={\%},
        every pin/.style={font=\scriptsize},
    ]{24.5/A, 25.2/B, 24.9/C, 25.4/D}
    \end{tikzpicture}
    \end{tabular}%
    }
    \end{minipage}%
    \hfill
    \begin{minipage}[t]{0.63\textwidth}
    \centering
    \resizebox{\textwidth}{!}{%
    \begin{tabular}{@{}c@{}}
    {\small\textbf{Modality Requirements}}\\[0.6em]
    \begin{tikzpicture}
    \pie[
        radius=1.6,
        text=pin,
        pin distance=0.4cm,
        color={modAV, modSound, modVisual},
        before number=\scriptsize,
        after number={\%},
        every pin/.style={font=\scriptsize},
        hide number,
    ]{99.0/{Both (A+V) -- 99.0\%}, 0.7/{}, 0.3/{}}
    \node[anchor=north west, font=\scriptsize] at (2.2, 0.4) {%
        \begin{tabular}{@{}ll@{}}
        \textcolor{modSound}{\rule{6pt}{6pt}} & Sound Only -- 0.7\% \\[2pt]
        \textcolor{modVisual}{\rule{6pt}{6pt}} & Visual Only -- 0.3\% \\
        \end{tabular}};
    \end{tikzpicture}
    \end{tabular}%
    }
    \end{minipage}
    \caption{Additional statistics for the AVRT-20K dataset, derived from the original AVQA annotations. \textbf{Top:} Question type distribution across training and validation splits closely mirrors the full AVQA dataset, with ``Which'' (45.2\%) and ``Come From'' (30.9\%) dominating, confirming that our dual-teacher filtering preserves the original distribution. \textbf{Bottom left:} Answer option distribution (training split) is balanced at ${\sim}25\%$ per option (A--D), mitigating selection bias. \textbf{Bottom right:} Modality requirements show that 99\% of questions require both audio and visual information to answer correctly, with only 1\% answerable from a single modality alone.}
    \label{fig:dataset_additional_stats}
    \end{figure*}
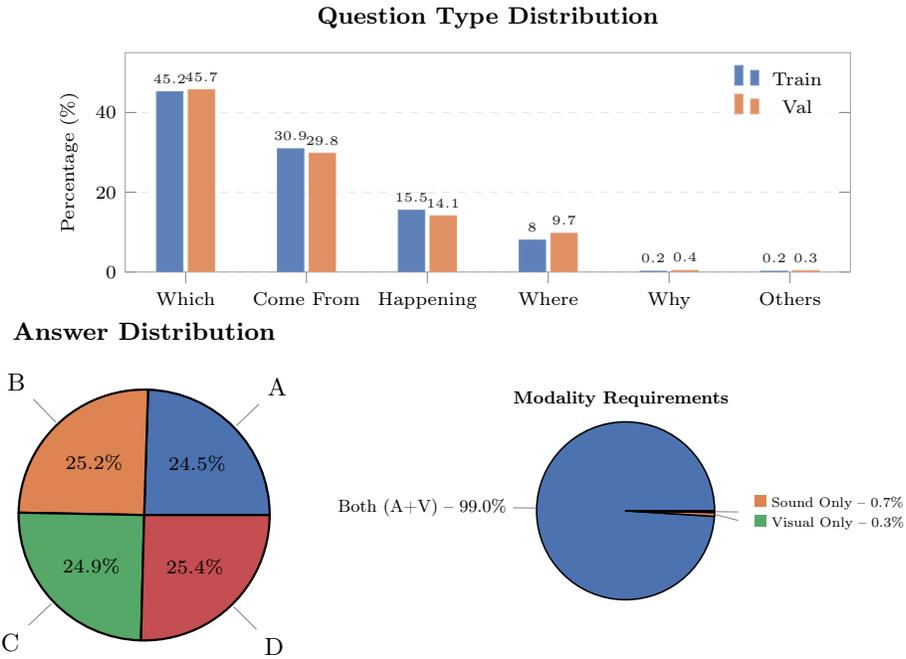

    Figure~\ref{fig:dataset_additional_stats} provides additional statistics for the AVRT-20K dataset. The question type distribution and modality requirements are derived directly from the original AVQA annotations. The question type distribution closely mirrors that of the full AVQA dataset, with ``Which'' questions being most prevalent (45.2\% in training), followed by ``Come From'' (30.9\%), ``Happening'' (15.5\%), and ``Where'' (8.0\%), confirming that our filtered subset preserves the original distribution. The dataset has balanced answer choices with each option (A--D) appearing roughly 25\% of the time, which is critical for mitigating selection bias in LLM training data. According to the AVQA modality annotations, nearly all questions (99\%) require both audio and visual information to answer correctly, with only 1\% being answerable using a single modality.

\section{Limitations and Future Work} \label{app:limitations_future}

One limitation of AVRT is that the quality of the generated multimodal traces is bounded by the factual reliability of the individual single-modality teachers. Our hallucination-source analysis suggests that most trace errors originate in the teachers rather than in the merger, and the pipeline may underperform if stronger unimodal teachers are not available for a given modality. Additionally, our SFT data is derived from a single source domain (AVQA), which may constrain the diversity of reasoning patterns seen during training. The generality of the approach to other modality combinations also remains to be established experimentally. Future work could address the teacher-quality bottleneck through uncertainty-aware merging and explicit trace verification, apply the pipeline to more diverse datasets to characterize SFT data scaling, and extend the framework to additional modality combinations, such as incorporating a speech-specific teacher for three-way merging. Iterative self-improvement, where the trained student serves as a teacher for subsequent rounds of trace generation, could also bootstrap higher-quality traces, since the student already outperforms individual teachers on several benchmarks.

\end{document}